\documentclass[3p, final]{elsarticle}

\usepackage{lineno,hyperref}
\usepackage{graphicx}
\usepackage{amsmath} 
\usepackage[FIGBOTCAP]{subfigure}
\usepackage{multirow}
\usepackage{lscape}

\journal{Journal}
\begin{document}

\begin{frontmatter}

\title{Optimal Compositions using Unconventional Modular Library for Customized Manipulators}%\tnoteref{mytitlenote}}

\author[oldAddress,currentAddress]{Anubhav Dogra}%\corref{mycorrespondingauthor}%\fnref{myfootnote}}
\date{}

\ead{anubhav.dogra@warwick.ac.uk, anubhav.dogra@hotmail.com}

\author[oldAddress]{Srikant Sekhar Padhee}
\author[oldAddress]{Ekta Singla}

\address[oldAddress]{Indian Institute of Technology Ropar, Punjab, India, 140001}
\address[currentAddress]{Present Address: University of Warwick, Coventry, United Kingdom}

\begin{abstract}
This paper presents an optimization approach for generating custom manipulator configurations using a proposed unconventional modular library. An end-to-end solution is presented in which the resulting optimal models of the modular compositions can be integrated directly with the Robot Operating System platform. The approach utilizes an unconventional modular library, which is adaptable to a wide range of parameters for customization including non-parallel and non-perpendicular joint axes, and the unified modeling technique for getting the custom modular configurations. The single objective function optimization problem is formulated based upon the discrete parameters of reconfiguration depending upon the available modular library such as, number of joint modules, skew-twist angle, intersecting-twist angle, connection ports of the module, module size, modular sub-assembly unit and curved links. Two case studies, including an application to the agricultural vertical farms, are presented to validate the results.
\end{abstract}


\end{frontmatter}

%\linenumbers
\section{Introduction}
New trends are heading towards the customization of the manipulator configurations to meet the rising demands of the custom products and services~\cite{bi2020framework}. The custom configurations are generally designed as per the given set of tasks to be performed in a given environment, however, the challenge is to design the configurations able to work in the cluttered work-spaces~\cite{singh2018modular}. The standard configurations available in the market may not able to work in a given new environment, and manufacturing a new fixed custom configuration is not an effective practice considering cost and time. To avoid the cycle of designing and re-manufacturing the systems from scratch, modularity and reconfigurability provides cost-effective solutions. Considering this, design of the modules for reconfigurable manipulators is presented by various researchers, such as, in~\cite{althoff2019effortless,Modman2020,hong2017joint, valsamos2014kinematic, acaccia2008modular, chen1997kinematic}. Lately, unconventional designs have been proposed which are able to adapt to the unconventional twist parameters required to perform the complex tasks~\cite{singh2018modular,brandstotter2018task,stravopodis2020rectilinear,dograJMD2021}. The term unconventional configurations mean that the configurations are having non-parallel and non-perpendicular adjacent joint axes. The use of non-parallel and non-perpendicular twist angles expands the design domain for the generation of customized manipulator configurations, which can work even in the cluttered environments where the conventional manipulators may fail to perform.

For the task-based configuration synthesis of the modular manipulators, there are mainly two techniques followed in literature considering various manipulator performance parameters. The first is the optimization of the robotic parameters, and then the modules are used to adapt to those parameters. The second is, to configure and optimize the modular compositions using the given module set. The optimization problems are being solved using deterministic or stochastic methods depending upon the type of problems targeted. A Denavit-Hartenberg (DH) parameters based approach can be seen in~\cite{singh2018modular}, in which the simulated-annealing based optimization method is used with heterogeneous modules having adaptive twist unit, considering varying Degrees of Freedom (DoF), task points and the cluttered workspaces. Similar method is used in~\cite{patel2015task} for the optimization of DH-parameters of a 6-DoF manipulator for a given set of tasks considering joint constraints, best kinematic performances and least power requirement. A memetic algorithm is presented in~\cite{tabandeh2016memetic} that combines both GA and local search algorithms, for task-based optimization of modular manipulators. A non-linear programming based approach is addressed in~\cite{whitman2018task} for the optimization of the robot design, using \textit{HEBI} actuators, along with the motion planning and elimination of the Degrees of Freedom (DoF). Same actuators are used in~\cite{campos2019task}, and proposed the task-based design for $2-4$ DoF systems only.

The technique based upon the enumeration of the modular compositions is used majorly to check the number of possible combinations and then optimizing the composition for given task~\cite{chen1998enumerating}. This is further combined with the dimensional synthesis, by using continuous and discrete variable at once and is solved for reachable errors and joint torques~\cite{bi2001concurrent}. A bi-level GA problem is presented in~\cite{chung1997task,chocron2008evolutionary} by generating modular compositions for each task, optimizing link lengths and avoiding obstacles. Enumerating the modular combinations and the unfeasible composition elimination method is presented in~\cite{Icer2016} for a given task considering path planning in task space and obstacles in the environment. Similar work is followed in ~\cite{icer2017evolutionary} using evolutionary approach. A 6-DoF unconventional configuration with pseudo joints is presented in~\cite{valsamos2014kinematic, moulianitis2016task}. The optimization problem is solved using GA with real coded chromosomes for best task dexterity indices at given task points and different types of paths. Optimization of 6-DoF manipulator for different paths using joint modules and curved links is proposed in~\cite{brandstotter2015curved}. The observations from the literature are as follows.

\begin{enumerate}
	\item Not much is explored in handling the non-parallel and non-perpendicular jointed configurations for developing customized manipulator to perform specific set of tasks in a given cluttered environment. This is interesting, as it expands the design space while synthesizing the robotic configurations.

	\item Enumerating the number of possible compositions becomes very large and difficult to handle when the design parameters available are large in number, as proposed in this paper. Also, the DoF of the configuration is generally kept fixed or handled through sequential addition or elimination methods in the optimization problem.

	\item Providing an end-to-end solution, from the task-based planning to the execution of the motion planning in a given environment, for the developed unconventional configurations need further exploration.

\end{enumerate} 
This paper proposes a new approach based upon the optimal modular compositions for the synthesis of the robotic configurations, which needs to perform a given set of tasks in a given cluttered workspace. Novelty of this work is presented through the generation of the modular compositions for the realization of non-parallel and non-jointed configurations, $n-$DoF, for the given cluttered environments. The approach utilizes the proposed modular library, named as \textit{MOIRs-MARK-2}~\cite{dograJMD2021}, and methodologies of unified modeling for unconventional configuration~\cite{dograUnified2022} directly for synthesizing the optimal modular compositions. DoF is considered as a design variable in the optimization problem, and handles reachability, joint torque limits, inverse kinematics, collision avoidance and motion planning of the unconventional configurations. The output model of the optimization can be directly used to on-the-fly development of a digital-twin in the ROS platform, and motion planning and execution can be done through ROS controllers.

The paper is organized as follows. Section~\ref{sec:problem} states the overall problem structure and the solution approach. Section~\ref{sec:modules} discusses the design of the modular architecture and the features. Section~\ref{sec:unified} defines the methodology for the unified and automatic modeling of the modular configurations. Approach for the modular composition optimization is presented in section~\ref{sec:approach}. A few case studies to validate the approach is presented in section~\ref{sec:results} followed by conclusion of the work in section~\ref{sec:conclusion}.
\section{Problem statement}\label{sec:problem}
The Problem is defined as $-$ Given the \textit{Task Space Locations} (\textit{TSLs}) and the cluttered environment, search a composition of the modules such that the configuration defined from the modular composition is able to reach all TSLs, plan the motion between the TSLs while avoiding collisions, and possesses minimum joint torques. The configuration is to be realized using the designed modular library which is unconventional in nature and the optimization problem takes care of the number of DoF, joint torque limits of the modules, inverse kinematics of the unconventional configurations for reachability, and motion planning while avoiding the obstacles in the cluttered environment. 

The task-based design approach based upon the optimization of the DH-parameters of `$n-$'DoF modular configuration followed by the adaptable modules to realize the optimum configurations is presented by the authors in~\cite{singh2018modular}. This paper is focused on proposing the reverse methodology, as to use the available modular data$-$ geometrical, inertial, meshes, assembly rules, reconfigurability, adaptable twist parameters $-$ for generating the optimal modular compositions. The problem is formulated with discrete variables depending upon the parametric reconfiguration of the modules such as, number of joint modules, skew-twist angle, intersecting-twist angle, connection port of the module, module weight, modular sub-assembly unit and link type. The methodology can be implemented to generate both conventional and unconventional modular compositions, and even for the modular libraries proposed by the various researchers in this domain.  
For this, brief descriptions of the proposed modular library and the unified modeling technique for the automatic generation of robot models~\cite{dograUnified2022} are highlighted in section~\ref{sec:modules} and \ref{sec:unified}, which are important for this paper. 

\section{Unconventional modular library}\label{sec:modules}
\begin{figure}[]
	\centering
	
	\subfigure[]{\includegraphics[width=1.55in]{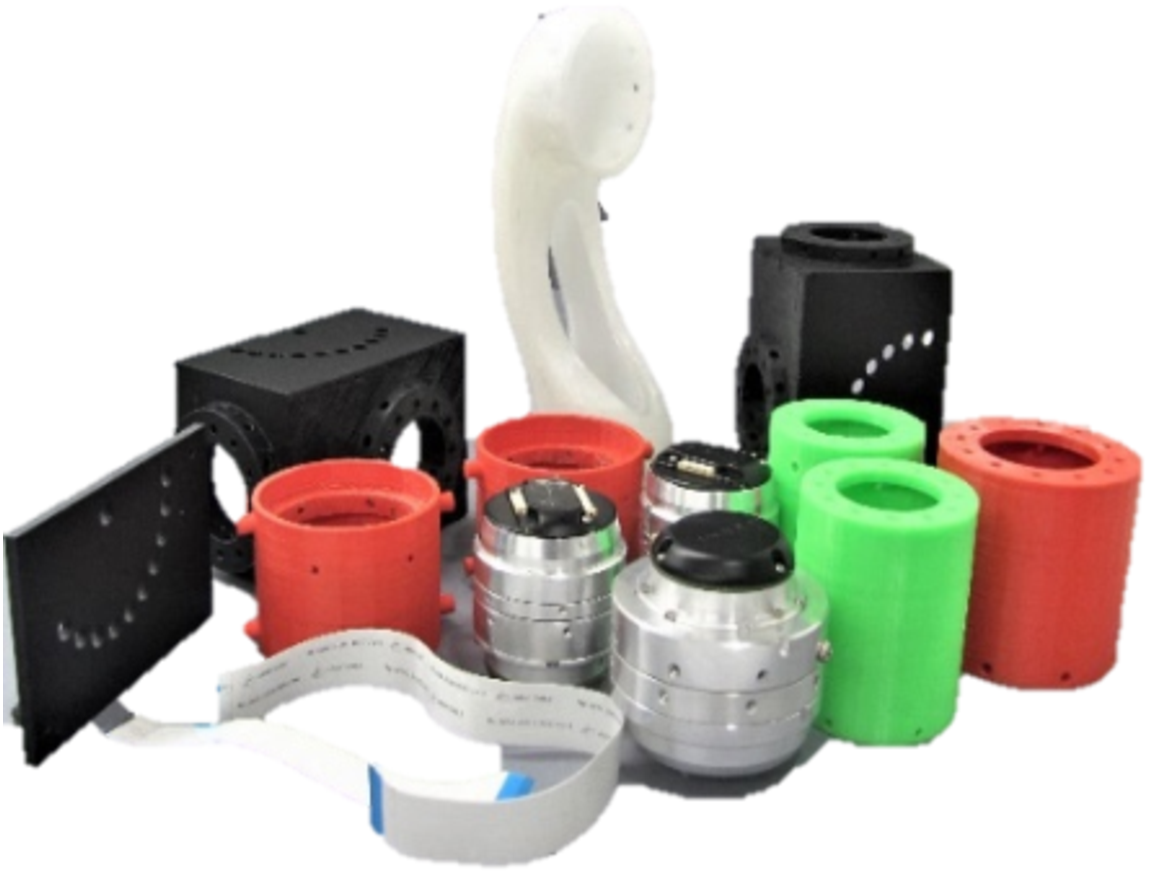}
	}~
	\subfigure[]{\includegraphics[width=1.55in]{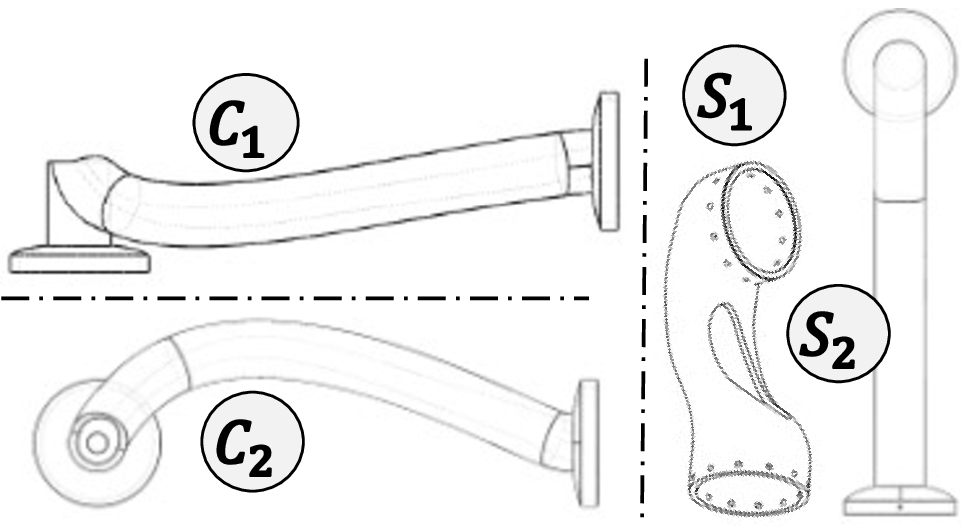}
	}~
	\subfigure[]{\includegraphics[width=2.1in]{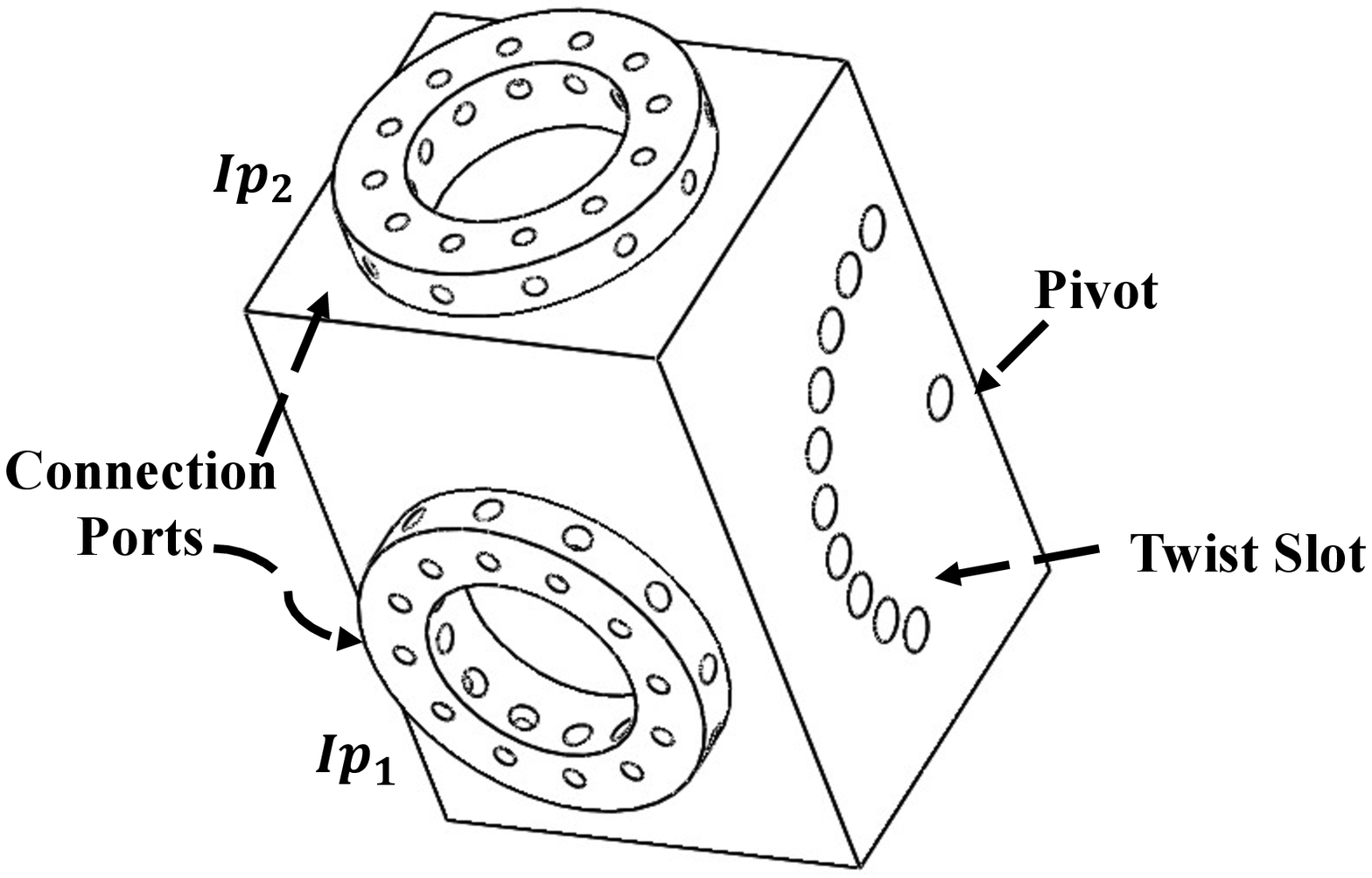}}
	~
	\subfigure[]{\includegraphics[width=0.65in]{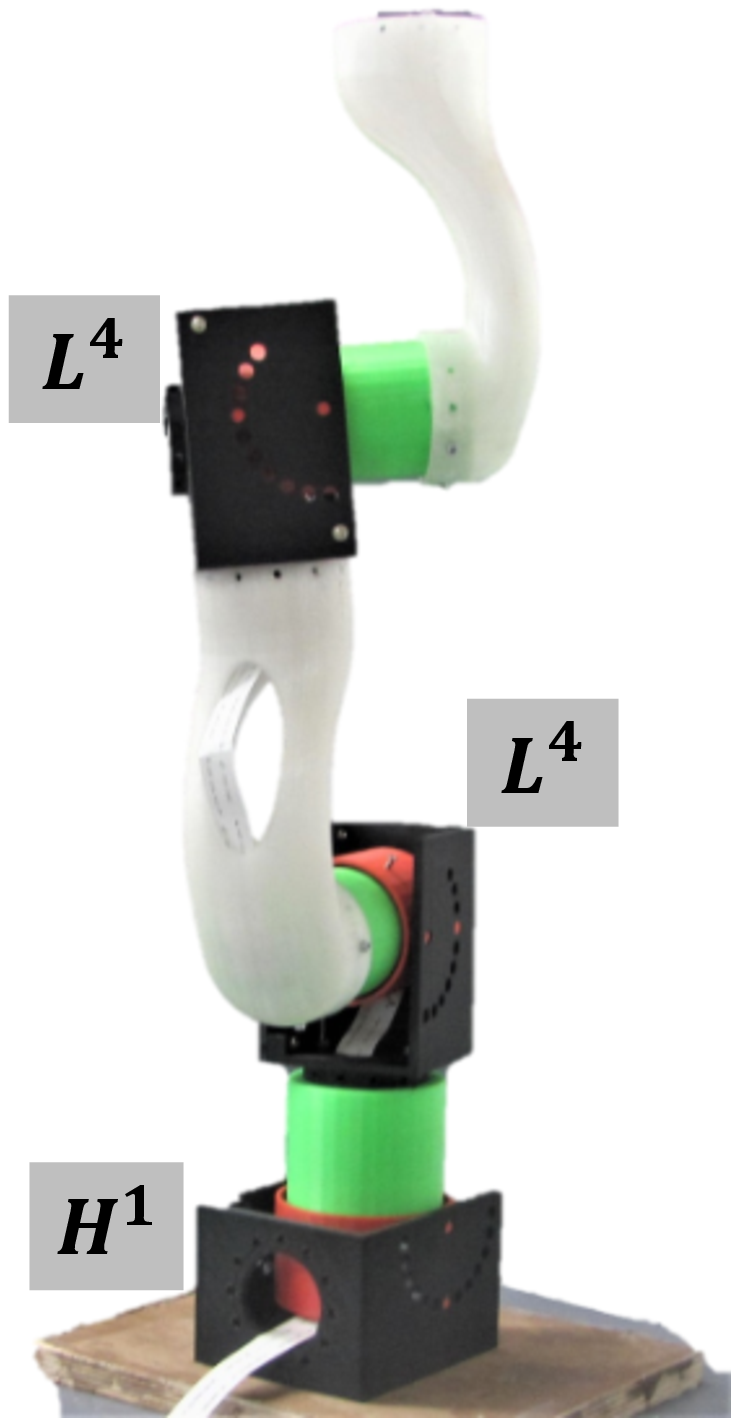}}
	\caption{(a) 3D-printed Modular library designed to assemble customized modular and reconfigurable manipulators~\cite{dograJMD2021}. (b) Link modules. (c) Twist adjusting unit.  (d) Exemplary 3-DoF modular configuration prototype, fabricated using a 3D printer with poly-lactic acid material.}
	\label{fig:joint_unit}
\end{figure}
A modular library, as shown in Fig.~\ref{fig:joint_unit}, consists of 1-DoF rotary joint modules and link modules, named as \textit{MOIRs' Mark-2}, to realize any custom manipulator configuration and is adaptable to a given set of robotic parameters. The detailed analysis of the modular library based upon the reconfigurability, types of configurations possible and assembly rules are proposed in authors work~\cite{dograJMD2021}. The modules are designed utilizing the optimal planning and design strategies~\cite{dograRobotica2021, dograJMD2022}, with respect to minimal dynamic torques. A joint module is an assembly of $3$ components as: unconventional twist unit, actuator and the actuator casings. Unconventional twist unit, as shown in Fig.~\ref{fig:joint_unit}(c) is designed to incorporate the twist angles adjustment between the two frames. The twist adjustment can be done in two ways. First is the angle between the two adjacent intersecting joint axes, and the second is an angle between two adjacent skew joint axes and about the common normal. The twist unit has a discontinuous semi-circular slot with resolution of $15^\circ$ ranges from $0$ to $+90^\circ$ and $-45^\circ$ in both clockwise and anticlockwise direction, as shown in Fig.~\ref{fig:joint_unit}(b). This slot is used to adjust the intersecting-twist angle. The skew-twist is adjusted using, two connection ports on the twist unit, as shown in Fig.~\ref{fig:joint_unit}(b).

\textit{KA-Series} actuators from \textit{Kinova}~\cite{kinova2019} are used for the joint modules, and are named Heavy (H) and the Light (L) modules based upon the two variants, named as $KA-75+$ and $KA-58$. Specifications of these actuators are provided in Table~\ref{tab:actuator_specs}. Here, $\epsilon$ denotes the number of modules which can be carried by the same type of modules when assembled~\cite{dograJMD2021}. Link modules are designed based upon the optimization strategies of \textit{Architecture Prominent Sectioning} $-k$ for minimum joint torques, as proposed by the authors in~\cite{dograJMD2022,dograRobotica2021}. The links are designed as of both conventional ($S_1, S_2$) and unconventional ($C_1, C_2$) type as shown in Fig.~\ref{fig:joint_unit}(b). The connection ports of the link are perpendicular to each other so as to assemble with the input and the output of the joint module respectively. 
\begin{table}[]
	\caption{Technical Specifications of Actuators}
	\begin{center}%\centering
		\label{tab:actuator_specs}
			\begin{tabular}{@{\extracolsep{\fill}}lcc}
				\hline
				Quantity & KA-75+ (H) & KA-58 (L) \\
				\hline
				$W~(kg)$ & 0.57 & 0.357 \\
				%				\hline
				$RPM$& 12.2 & 20.3 \\
				%				\hline
				$\tau_{nom}~ (Nm)$& 12 &  3.6 \\
				%				\hline
				$\tau_{max}~ (Nm)$& 30.5 &  6.8 \\	
				%				\hline
				$\epsilon$&3&3\\	
				\hline
			\end{tabular}
	\end{center}
\end{table}
\section{Unified modeling of modular configurations}\label{sec:unified}
Whenever the modules are assembled together to develop $n-$DoF manipulator, or if the reconfiguration is done by changing the twist parameters, links, assembling with any of the connection ports, etc., the whole kinematics and dynamics of the modular configuration changes. To formulate the models automatically, the unified modeling is proposed~\cite{dograUnified2022}. Unified modeling in this paper is about the automatic generation of the Unified Robot Description Format (URDF) models for the serial manipulators composed of the components of the modular library. The benefit of generating the URDF files is that, these models can be integrated directly in the robotic tool-boxes such as in MATLAB\textsuperscript{\textregistered{}} and in ROS, thus saves time from re-writing the necessary codes for the computations of kinematics, dynamics, collision avoidance, motion planning, etc. The developed virtual models are used not only to simulate the design in a given environment but also to execute the developed configurations using Robot Operating System (ROS) based platform.

\begin{figure} 
	\centering
	\subfigure[]{\includegraphics[width=0.95in]{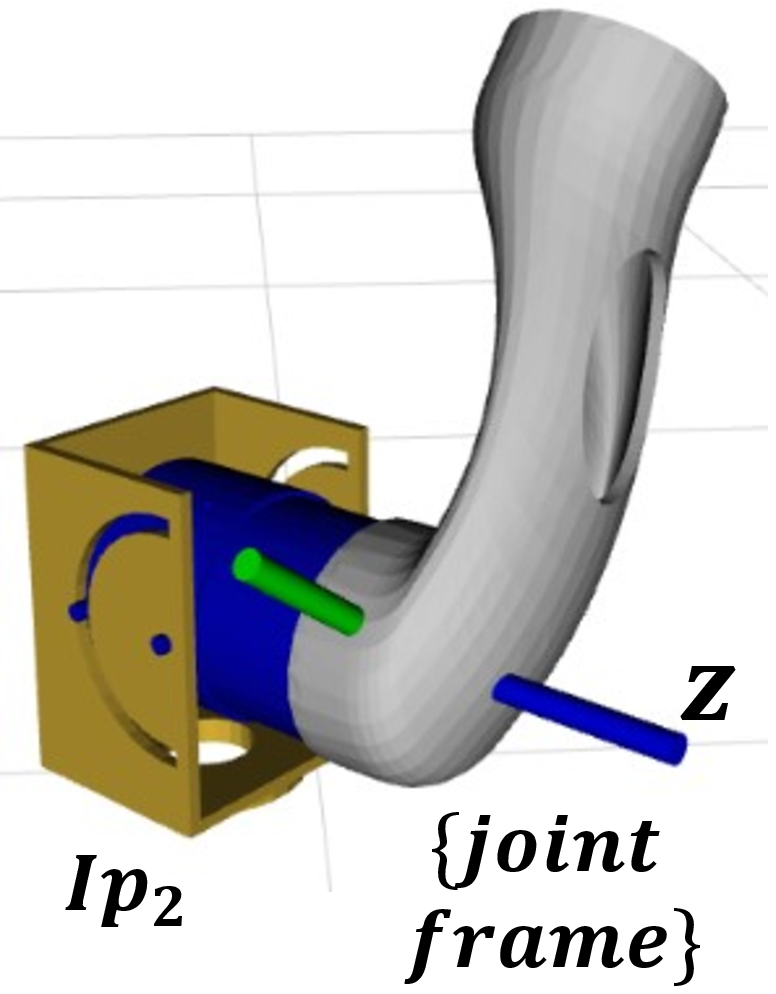}}
	~~~~~
	\subfigure[]{\includegraphics[width=0.95in]{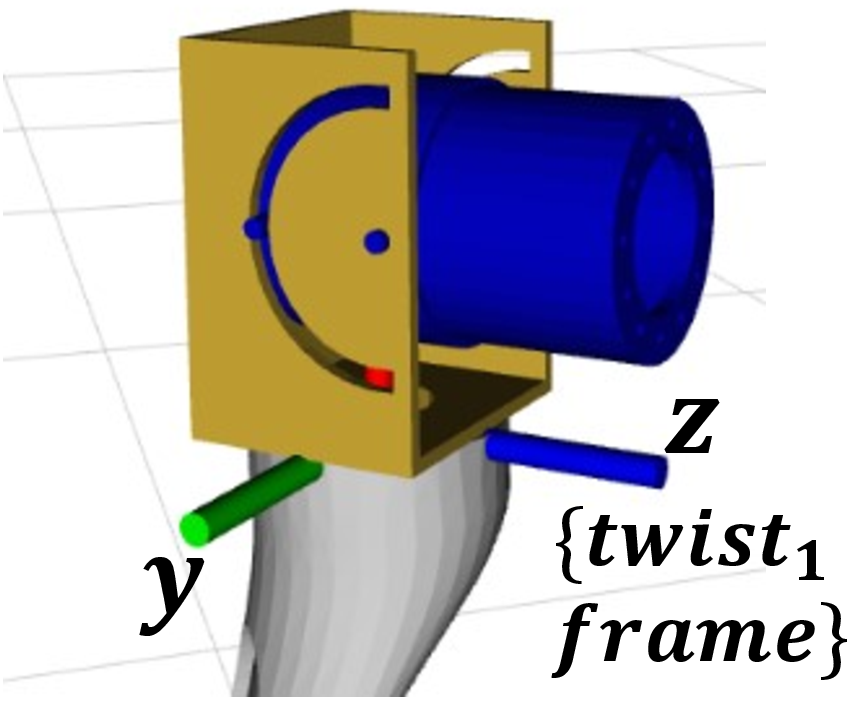}}
	~~~~~
	\subfigure[]{\includegraphics[width=1.05in]{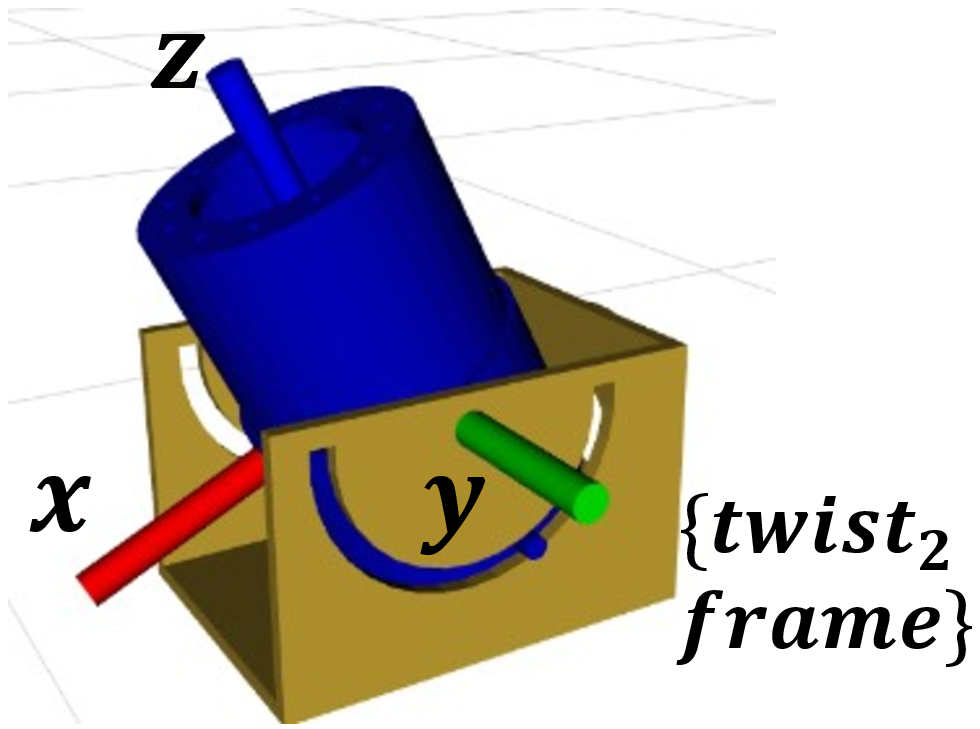}}
	\\
	\subfigure[]{\includegraphics[width=0.75in]{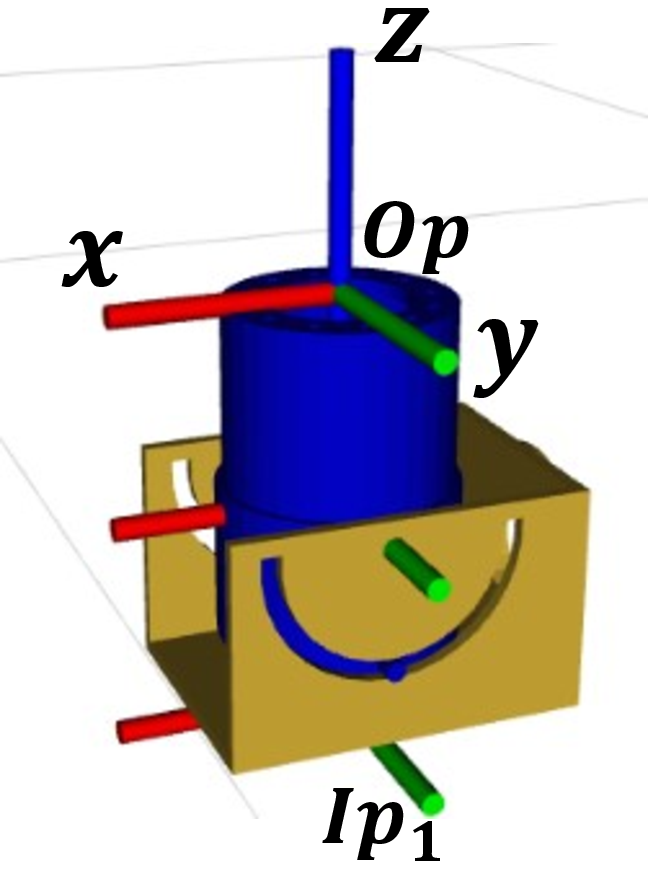}
	}
	~~~~~
	\subfigure[]{\includegraphics[width=1.05in]{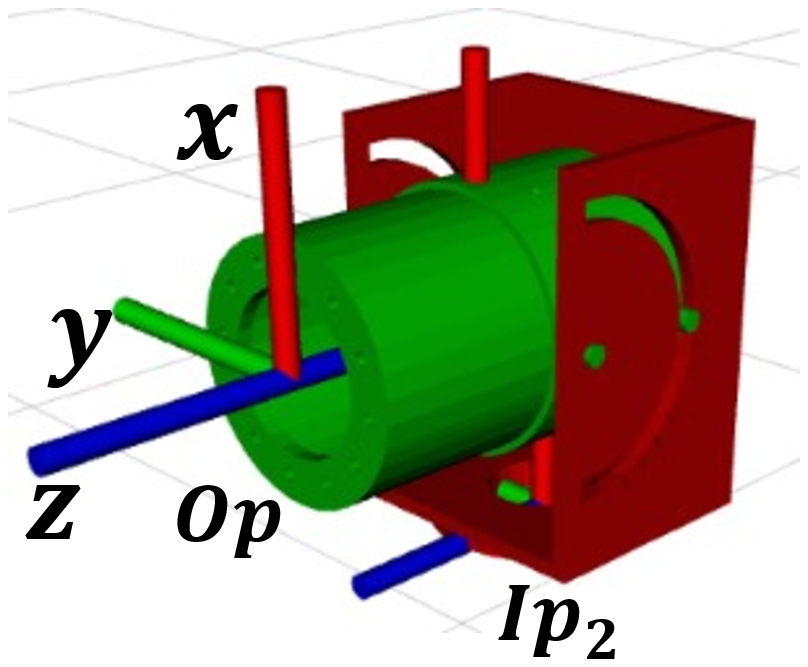}
	}
	~~~~~
	\subfigure[]{\includegraphics[width=1.25in]{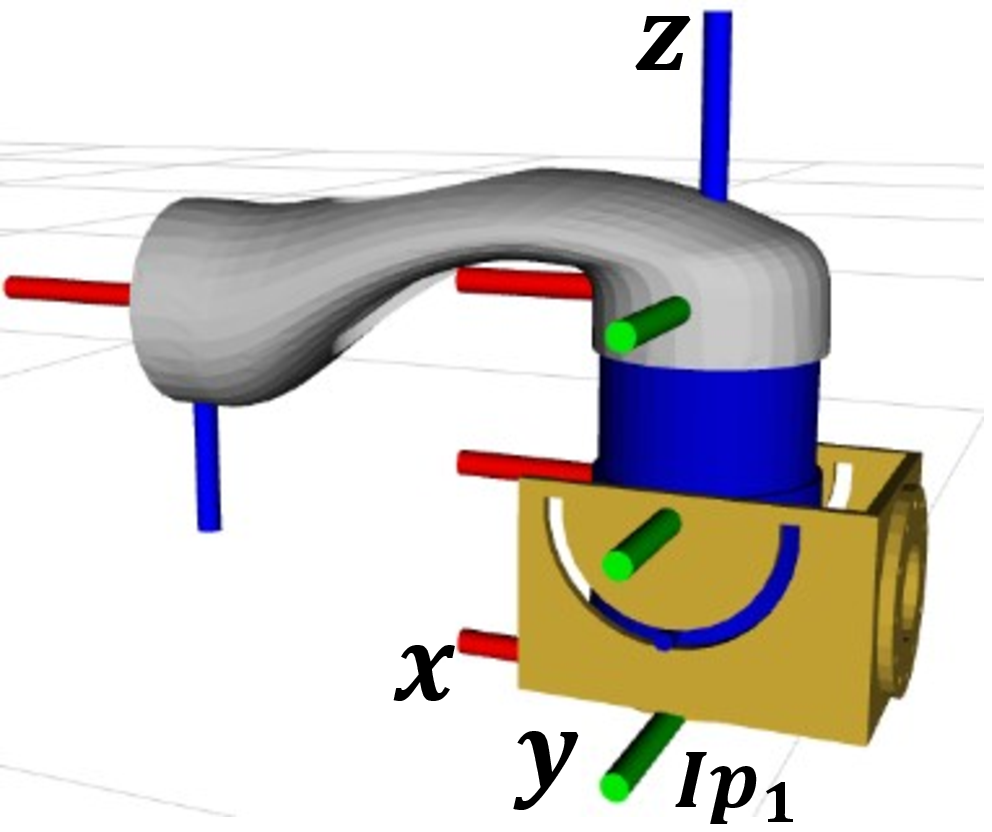}
	}
	\caption{$3$ Frames associated to modular components and modular units, (a) Blue axis represents the z-axis of joint frame and is located on the actuator axis (b) Red axis represents the x-axis of twist$_1$ frame located at the connection port (c) Green axis represents the y-axis of the twist$_2$ frame located inside the twist adjustment module, (d) Modular unit type $H^1$ or $L^1$, (e) Modular unit type $H^2$ or $L^2$, (f) Modular unit type $H^3$ or $L^3$, and $H^4$ or $L^4$ based upon the connection port to be assembled with.}
	\label{fig:frames}
\end{figure}

For the unified modeling of the modular compositions, first the structural representation is proposed to identify a generated modular composition. Each joint module is having $3$ connection ports in total through which joint or link modules can be assembled. Two of them are input connection ports ($Ip_1,~Ip_2$) and the other is an output connection port ($Op$), as shown in Fig.~\ref{fig:frames}. The $(k-1)^{th}$ module is connected to the ${k^{th}}$ module at the input connections ports ($Ip$), and the ${(k+1)^{th}}$ module is connected to the ${k^{th}}$ module through output connection port ($Op$). A set of rules for the physical assembly of any required configuration assembly are prepared as follows.
\begin{enumerate}
	\item The first module to be used in the composition is always the Heavy (H) variant and the last module is always the Light (L) variant, if the DoF is large than 3.
	\item Connect $H_{i}$($Op$)~$\rightarrow$ $H_{i+1}$ ($Ip_{1_{i+1}}$) or $H_{i+1}$ ($Ip_{2_{i+1}}$) or $H_{i}$($Op$) $\rightarrow$ $L_{i+1}$ ($Ip_{1_{i+1}}$) or $L_{i+1}$ ($Ip_{2_{i+1}}$).
	\item $L_{i}$($Op$) $\rightarrow$ $L_{i+1}$ ($Ip_{1_{i+1}})$ or $L_{i+1}$($Ip_{2_{i+1}}$).
	\item Refer $\epsilon$ for maximum number of each module from Table~\ref{tab:actuator_specs}.
\end{enumerate}
Here $i=1:n$, with \textit{n} as number of DoF. The guidelines are defined for the systematic generation of the modular composition, and not to jumble the variants (H or L) of the modules in a composition. Therefore, through point 1, it is suggested to start with the H module. If the composition is to be started with the L module, then the H module is not recommended to be assemble after the L module. Any \textit{n-DoF} manipulator will be the sequence of H and L modules, carrying a given range of payloads, with considerations of the above mentioned rules.

For the reconfiguration to be performed within the modules, the three orthonormal frames are associated to the modular components, as shown in Fig.~\ref{fig:frames}. In this paper, red, green and blue axes represents x, y and z axes. Joint frame is a rotating frame, with variable as joint angle, about its own $z-$axis (blue), as shown in Fig.~\ref{fig:frames}(a). It is also an attaching frame for the connection of the modules with the $Op$ port of the joint module. Twist$_1$ frame connects previous modules with the $Ip_1$ port of the joint module. Adjustment of the skew-twist is done by rotating the joint module about the $x-$axis (red) of twist$_1$ frame, as shown in Fig.~\ref{fig:frames} (b). Twist$_2$ frame lies inside the joint module and is on the pivot axis with origin at its center. Intersecting-twist is adjusted by rotating the actuator casings about $y-$axis (green) of twist$_2$ frame as shown in Fig.~\ref{fig:frames} (c). Along with this, $4$ modular units (sub-assemblies) are proposed here for the easy assembly of the joint and the link modules, both for the software and for the hardware assemblies. The modular units are valid for both `H and `L' variant of the joint modules and can be enumerated as, `$H^k$' associated to Heavy (H) modules and `$L^k$' associated to Light (L) modules, $\forall ~k\in\{1:4\}$. 
\begin{enumerate}
	\item `$H^1$' or `$L^1$' is used when a joint module is to be connected through $Ip_1$, as shown in Fig.~\ref{fig:frames} (d).
	\item `$H^2$' or `$L^2$' represents the corresponding joint module connected through $Ip_2$, as shown in Fig.~\ref{fig:frames} (e).
	\item `$H^3$' or `$L^3$' is used when a joint module is connected through $Ip_1$ and it is having a link module attached at $Op$, as shown in Fig.~\ref{fig:frames} (f).
	\item `$H^4$' or `$L^4$' represents a joint  module connected through $Ip_2$ and having a link module at $Op$, as shown in Fig.~\ref{fig:frames} (a). 
\end{enumerate}
So finally, any modular composition will be represented by the sequence of $4$ modular units and the twist parameters are to be adjusted by the twist$_1$ and twist$_2$ frames. An algorithm is developed, which takes the modules type-sequence ($H^k$' or `$L^k$) along with the robotic parameters to automatically generate the desired virtual robot model, i.e., the URDF file. A URDF is written in an \textit{extensive markup language (xml)}, which consists of various \textit{xml} tags such as, joint tag, link tag, geometric file tag, inertial tag, collision tag, positional and orientational details about the joint and the link frames (in Cartesian coordinates and Euler angles respectively), etc. A database has been provided which contains the \textit{stl} models, with the information of geometrical and inertial data, of the components of the modular library to be used in the generation of URDF file. Therefore, in the optimization process, the URDFs are automatically generated with the given modular unit sequence and reconfiguration parameters which acts as the design variables of the formulation.     

\section{Task-based design approach}\label{sec:approach}
The approach refers to the synthesis of robotic manipulator configurations which are able to perform the given set of tasks in the given cluttered environments. The robotic configuration to be synthesized, is to be realized using the modular library. To formulate the problem, it is important identify the factors or the parameters which are playing role during the optimal synthesis. These factors are to be formulated as the objective function and the constraint equations in the optimization model. The objective function and the constraint functions are computed using the unified models of the modular compositions.
\begin{figure}[]
	\centering
	\includegraphics[width=2.5in]{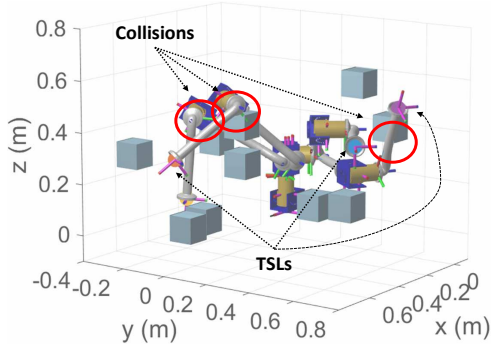}
	\caption{3-DoF configuration reaching the TSLs in a cluttered environment but is in collision at the three marked locations}
	\label{fig:collision_proof}
\end{figure}
The most important factor for the synthesis of any manipulator is its reachability for the Task Space Locations (TSLs) which has to be satisfied for a complete path. If the manipulator is not able to reach then there is no point of deploying that configuration on site. The reachability is also need to be assured with the collision avoidance in the environment of deployment. Figure~\ref{fig:collision_proof} shows a 3-DoF configuration in a cluttered environment able to reach on the TSLs. However, the configuration is in collision with the environment at the marked locations, as shown in Fig.~\ref{fig:collision_proof}, even for all possible inverse kinematic solutions. The composition is developed as of standard configuration, using 3 joint modules and 2 link modules. 

This marks the need of expanding the design variable space in optimization, especially for twist angle parameters in the configuration. It could be possible that the configuration can be synthesized with larger DoF while it also avoids the obstacle, but the target here is to get the configuration with minimum DoF and with minimum joint torques requirement. The other important factors that can be considered for task-based optimization problem are like energy consumption or joint torques, obstacle avoidance from the collision environment, assembly feasibilities, inverse kinematics, motion planning between the TSLs, manipulator performance in its workspace, etc. 

\subsection{Design variables}
For the configuration synthesis, it is proposed here to use the elements of modular library for each module and type, as briefed in section~\ref{sec:unified}. Therefore, the set of design variables ($x$), as given in Eq.~\ref{eq:DVs},  used in this optimization formulation are discrete in nature. The design variable are also shown in Table~\ref{tab:DVs}, and are as follows.
\begin{enumerate}
	\item \textbf{DoF ($n$)}: The DoF of the system in the optimization formulation is kept as design variable and is allowed to be varied within the range of $2-6$.
	\item \textbf{Joint Module Variant ($v_m$):} There are two variants of the joint module named as Heavy (H) and Light (L) based upon the size of the actuator used in it. The number of H or L modules to be used in sequence are to be decided using assembly rules and $\epsilon$ parameter, as discussed in section~\ref{sec:unified}. This is to be taken care in the optimization formulation.
	\item \textbf{Modular unit ($u_m$)}: For the assembly of the modular configurations, 4 modular units/sub-assemblies are proposed for each module type as $H^1$, $H^2$, $H^3$, $H^4$, and $L^1$, $L^2$, $L^3$, and $L^4$. The optimal modular composition will be the sequence of the these modular units.  
	\item \textbf{Link type ($l_m$)}: 4 type of links can be used in the optimization strategy, consist of 2 straight and 2 curved links. These are named as $S1,~ S2,~ C1$ and $C2$, as shown in Fig.~\ref{fig:joint_unit}(b)
	\item \textbf{Twist angle (\boldmath$\alpha$\unboldmath)}: There are two types of twists as type II and III, which can be adjusted through the mechanism presented in the modular joint unit. That is, using the connection ports and the adaptive twist unit which varies from $-45^\circ$ to $90^\circ$. The valid assemblies are only allowed to participate in the configuration synthesis, otherwise penalties are applied in the formulation.
\end{enumerate} 
\begin{equation}
	\mathbf{x}=[n, ~\{v_m, u_m, \alpha, l_m\}_1, ~\{v_m, u_m, \alpha, l_m\}_2,.....,~\{v_m, u_m, \alpha, l_m\}_{n}].
	\label{eq:DVs}
\end{equation}
\begin{table}[]
	\caption{Design Variables for Optimization Problem}
	\begin{center}%\centering
			\label{tab:DVs}
			\begin{tabular}{@{\extracolsep{\fill}}lcc}
				\hline
				Design variable type & Variants \\
				\hline
				%				\hline
				DoF ($n$)  & $2-6$\\
				%				\hline
				Module variant ($v_m$)& $H, ~L $\\
				%				\hline
				Modular Unit ($u_m$) & $H^k$ or $L^k$ $\forall ~k\in\{1:4\}$\\
				%				\hline
				Link type ($l_m$) & $S_1,~S_2, ~C_1, ~C_2$ \\
				%				\hline
				Twist angle ($\alpha$) & $[-45,-30,-15,0,15,30,45,60,75,90]$\\
				\hline
			\end{tabular}
	\end{center}
\end{table}
The DoF is also considered as one of the design variables in the optimization study, therefore, the total number of the design variables available during the optimization process would change for every iteration. The total number of design variables would be $4n+1$ for each candidate solution in the optimization algorithm. Geometrical and inertial parameters of each component along with the solid models are stored in the database to be used for computation of objective function and the constraint functions. Due to the use of discrete variables and varying number of design variables, genetic algorithm based optimization solver is used in this work using \textit{global optimization tool-box} of MATLAB\textsuperscript{\textregistered{}} R2021a.  
\subsection{Objective function}
The objective of the optimization problem is defined as to find a modular composition from the given set of modules, which can reach for all the TSLs, and possesses minimum joint torques for the manipulator configuration over all the TSLs. It could be possible that there are many possible configurations, with DoF ranging from $2-6$, that would be able to reach the target positions. Considering that the designer would like to have a configuration that would require minimum energy or the actuator efforts while moving in between the given TSLs. Therefore, the objective function in this is formulated to minimize the DoF of the modular composition which will lead to have minimum joint efforts requirement, and can be defined as sum of root mean square ($rms$) of joint torques of all the joints in the synthesizing configuration over
all the TSLs. The joint torque for the configuration can be computed using Euler-Lagrange equations of motion as in ~\cite{dograJMD2021}. This can be also be implemented through MATLAB\textsuperscript{\textregistered{}} subroutines named as \textit{inverseDynamics} and \textit{gravityTorque}.
\begin{eqnarray}\label{eq:rms}
	f(\textbf{x}) = \sum_{i=1}^{n} rms(\tau_{j}), ~\forall ~j \in \{1:N\},
\end{eqnarray}
where, $f(\textbf{x})$ is the objective function, $N$ is the number of TSLs and $n$ is number of joint modules (DoF).
\subsection{Constraint equations}
Modular composition to be generated during optimization has to satisfy the following constraints. 
\subsubsection{Reachability}
The end-effector of the generated configuration has to reach all the given TSLs. The reachability is the pose error norm between the end-effector pose and the TSL pose. The constraint is an equality type so that the error norm should be strictly satisfied. This might include both the position and the orientation error of the end-frame of the synthesizing robot with the TSL frame.
\begin{eqnarray}
	\mathbf{g}_1(\textbf{x}) = \|\mathbf{P}_{tsl} - \mathbf{P}_{eef}\|_j = 0.\\
	\forall ~j\in \{1:N\}\nonumber
\end{eqnarray} 
where, $\mathbf{P}_{tsl}$ defines the pose of TSL frame, and $\mathbf{P}_{eef}$ defines the pose of the end-effector frame of the modular composition. 
\subsubsection{Collision avoidance}
To avoid collision of the modular elements with the environment (obstacles), the minimum distance $(D)$ between the any two meshes is computed and is used for the collision check~\cite{2083}. The environment is assumed as the assemblage of primitive shapes (cuboid, sphere, cylinders) and are converted into the triangulated meshes. The geometrical data of the modules are retrieved from the \textit{stl} files and are also converted to the triangular meshes. The collision check is done among links (self collisions) and between the links and the obstacles (environment collision). Thus, non-linear inequality constraint can be written as
\begin{eqnarray}
	&\mathbf{g}_{2}(\textbf{x})=\delta-D_{pq}\leq 0;\\ &\forall~ p \in \{1:N_{mod}\}, ~q \in \{1:N_{obs}\}\nonumber
\end{eqnarray}  
where $\delta$ is the safety margin to avoid collision, $N_{mod}$ represents number of modular elements and $N_{obs}$ represents number of obstacles.
\subsubsection{Joint torque limits}
This constraint is used to limit the torque values of each joint of the configuration. $\epsilon$ is the parameter which defines how many number of modules of one kind can be carried by the same module itself, as given in Table~\ref{tab:actuator_specs}. This acts as a limit to the number of modules to be assembled in sequence keeping the account of module type used (H or L). This has been decided based on the study conducted considering the worst torque analysis of the joint torques, for varying range of the payloads and range of DoF, for each kind of module~\cite{dograJMD2021}. Therefore, for each modular joint and the type, the corresponding nominal joint torque limits of the actuators are used, as given in Table~\ref{tab:actuator_specs}.
\begin{eqnarray}
	\mathbf{g}_{3}(\textbf{x})=
	&\begin{cases}
		\tau_{ij} - 12 \leq  0 & \text{if } i \in H\\
		\tau_{ij} - 3.6 \leq  0    & \text{if } i \in L
	\end{cases}\\
	&\forall ~i\in \{1:n\}, j ~\in \{1:N\}.\nonumber
\end{eqnarray}

\subsection{Optimization process}
\begin{landscape}
	\begin{figure}
		\centering
		\includegraphics[width=\columnwidth]{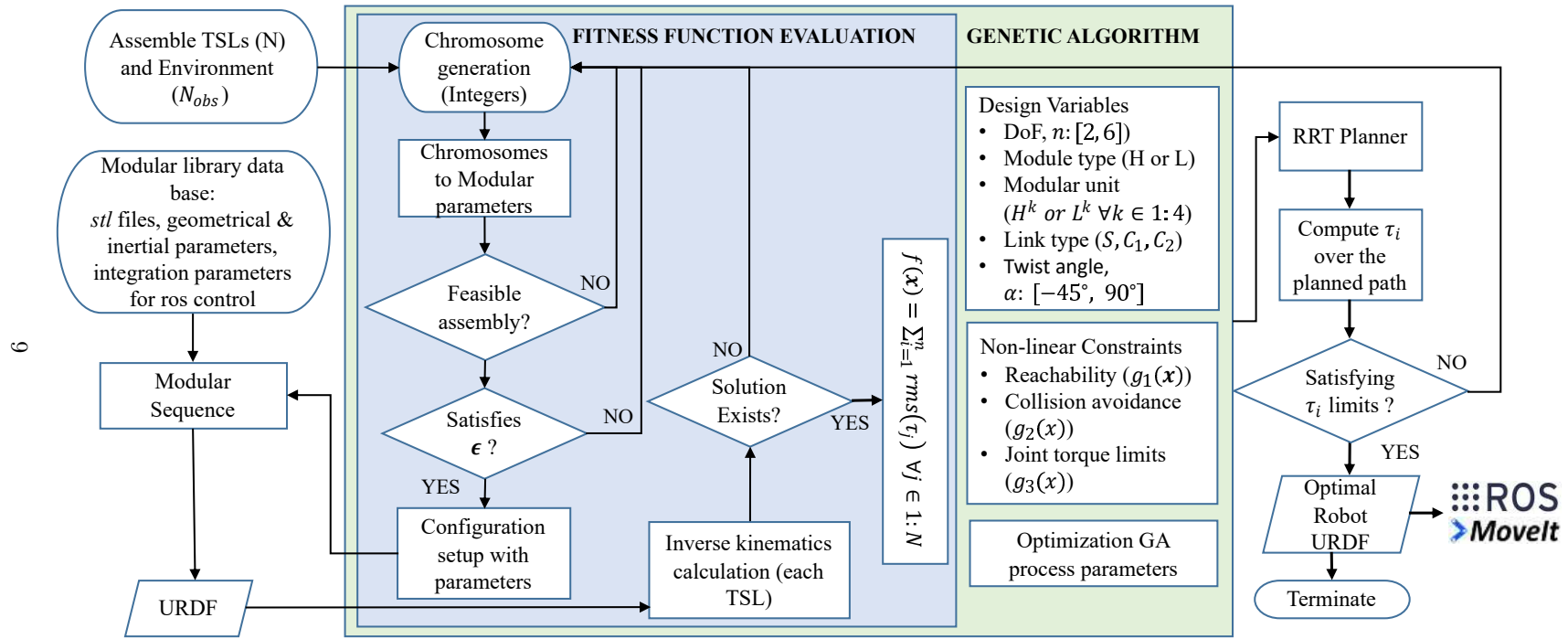}
		\caption{Proposed approach of the optimization is shown through a flow diagram.}
		\label{fig:GAflow}
	\end{figure}
\end{landscape}

Generally while formulating the problem, along with the robotic parameters, joint angles of the synthesizing configuration are often used as design variables. Due to the large search space of the joint angles and its nature of the variable type, which is continuous, joint angles are not considered as the design variables in the current approach. Therefore, in this work, inverse kinematics based solutions are proposed which computes the joint angles for all the TSLs but are not used as design variables. The proposed optimization strategy is briefed in Fig.~\ref{fig:GAflow}. The optimization problem is formulated with discrete variables and is solved using global optimization toolbox of MATLAB\textsuperscript{\textregistered{}} R2021a.
First the TSLs and the environment data is assembled and are passed as input for the optimization problem for which the manipulator is required to be customized. The robot base location is located at $(0, 0, 0)$ and all the other entities, such as TSLs and the workspace, are defined with respect to this frame of origin. The TSLs are given in terms of the Cartesian coordinates for position and Euler angles in \textit{`XYZ'} convention for orientation parameters. The environment data is given as primitive collision boxes using MATLAB\textsuperscript{\textregistered{}} robotics toolbox function - $collisionBox$. Then, inside GA, the generated chromosomes (discrete variables) are mapped into the modular parameters as per the design variables. The sequence of the modular elements are combined with the twist angles and is checked for the feasible assembly, and the $\epsilon$. After this, modular sequence is used to automatically generate the URDF file which represents the modular configuration. The inverse kinematics computation of the configuration is done through the numerical solutions using the MATLAB\textsuperscript{\textregistered{}} function - $inverseKinematics$ - for all the TSLs. The computed joint angles are then used to compute the joint torque, for every joint and each TSL, to compute the objective function. The constraints, bounds and optimization parameters are handled through subroutines of the GA.

The output file from the GA can be then checked for the path planning between the TSLs using \textit{RRT-connect}~\cite{kuffner2000rrt} path planner and then the joint torque limits can be verified for the entire path. The final optimal configuration, converted to URDF file, can directly be used in the Robotic Operating System (ROS) for the motion control of assembled modular manipulator, as shown in Fig.~\ref{fig:GAflow}
\section{Results}\label{sec:results}
To demonstrate the proposed strategy for design synthesis of the unconventional modular manipulators, two case studies are presented with two different type of environments and the corresponding TSLs. Case I is to demonstrate the optimization strategy for a highly cluttered environment by considering obstacles made of primitive shape distributed randomly in 3D space. Case II is focused on a realistic environment of agricultural vertical farm for which a custom configuration is required to accomplish the tasks of transplantation. The results have validated the utilization of the proposed modular library and the unified methodologies for the optimal generation of modular compositions. 

\begin{figure}[ht]
	\centering
	\subfigure[]{\includegraphics[width=2in]{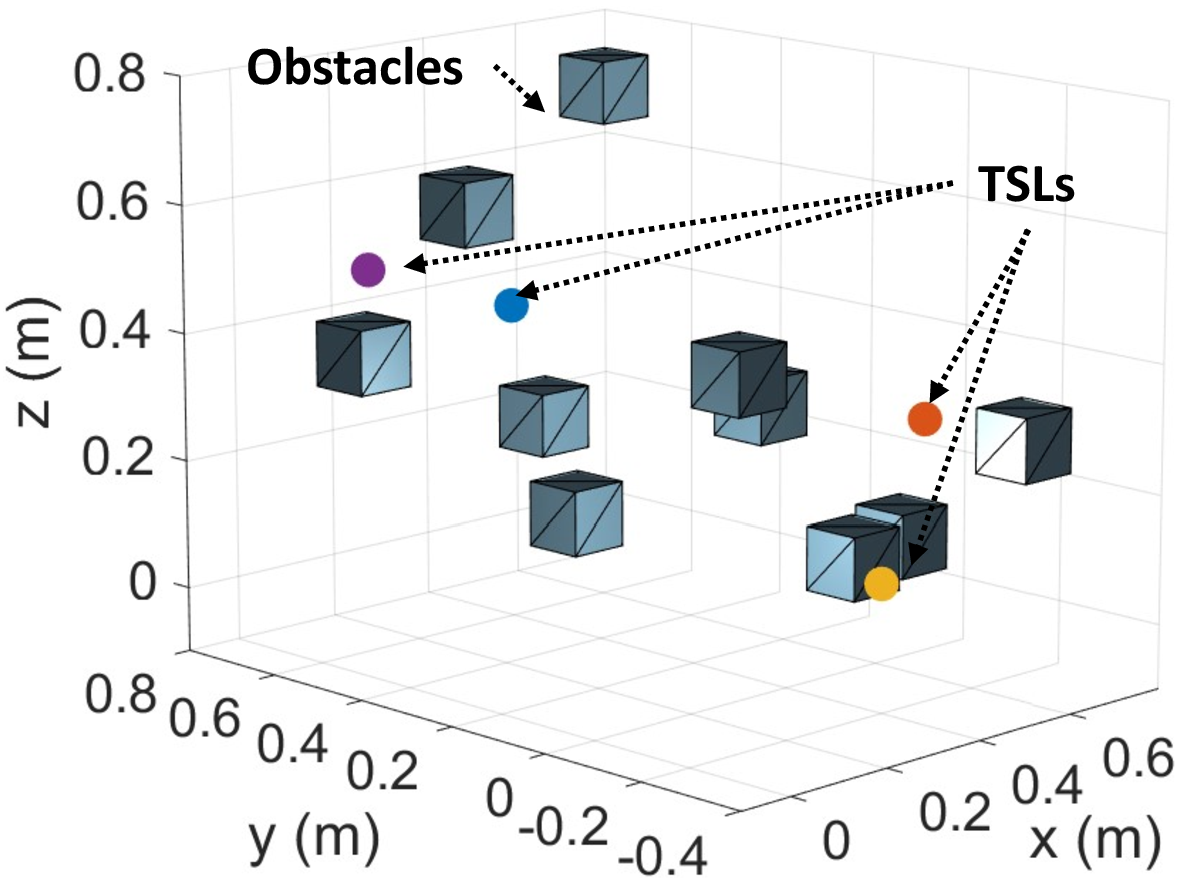}}
	~
	\subfigure[]{\includegraphics[width=1in]{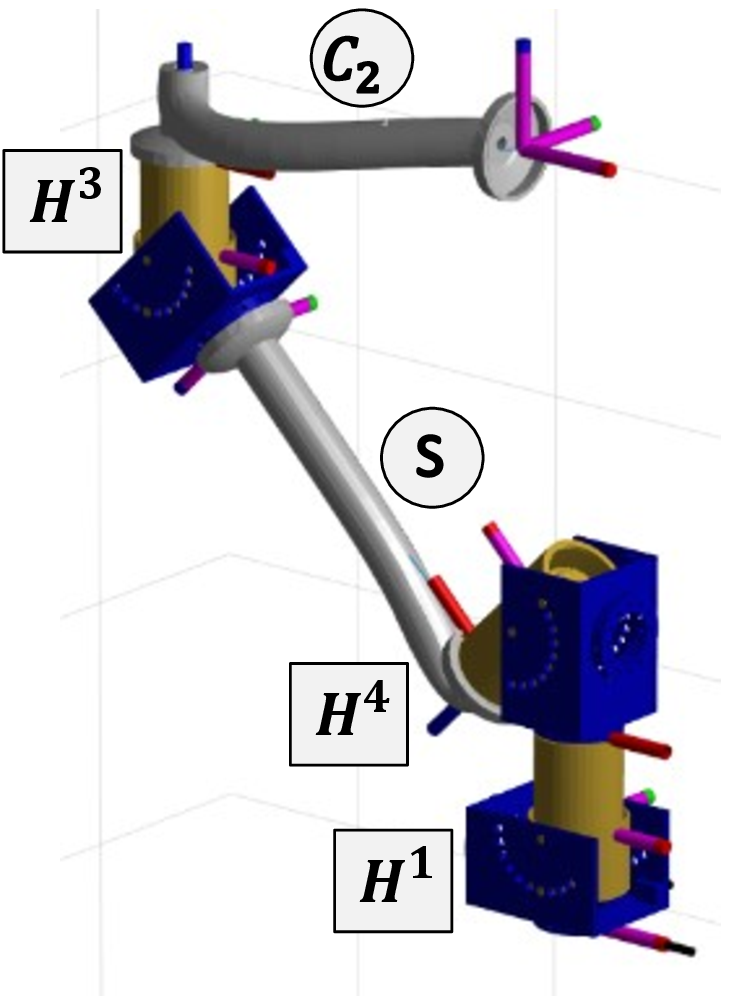}}
	~
	\subfigure[]{\includegraphics[width=2in]{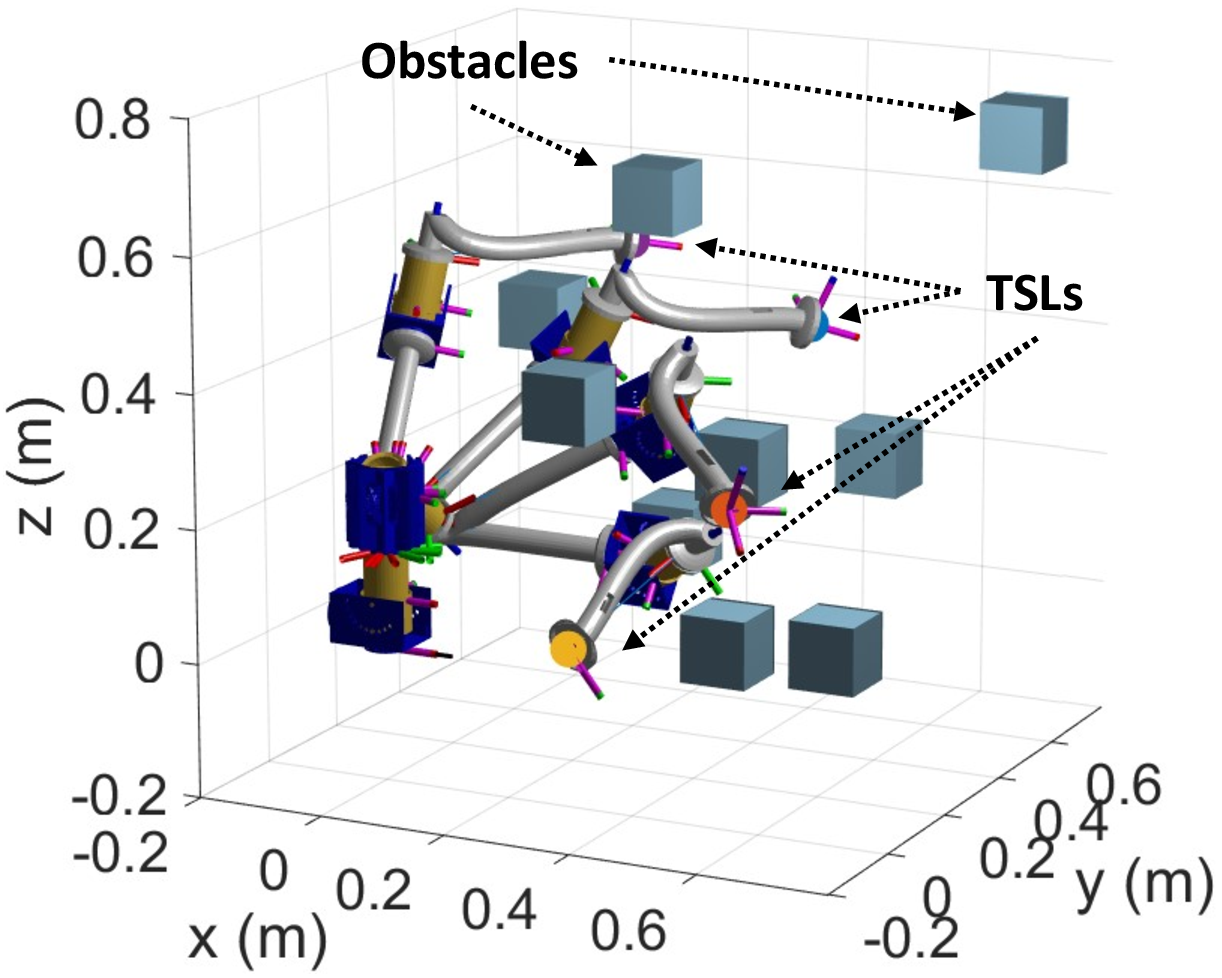}}
	~
	\subfigure[]{\includegraphics[width=2.1in]{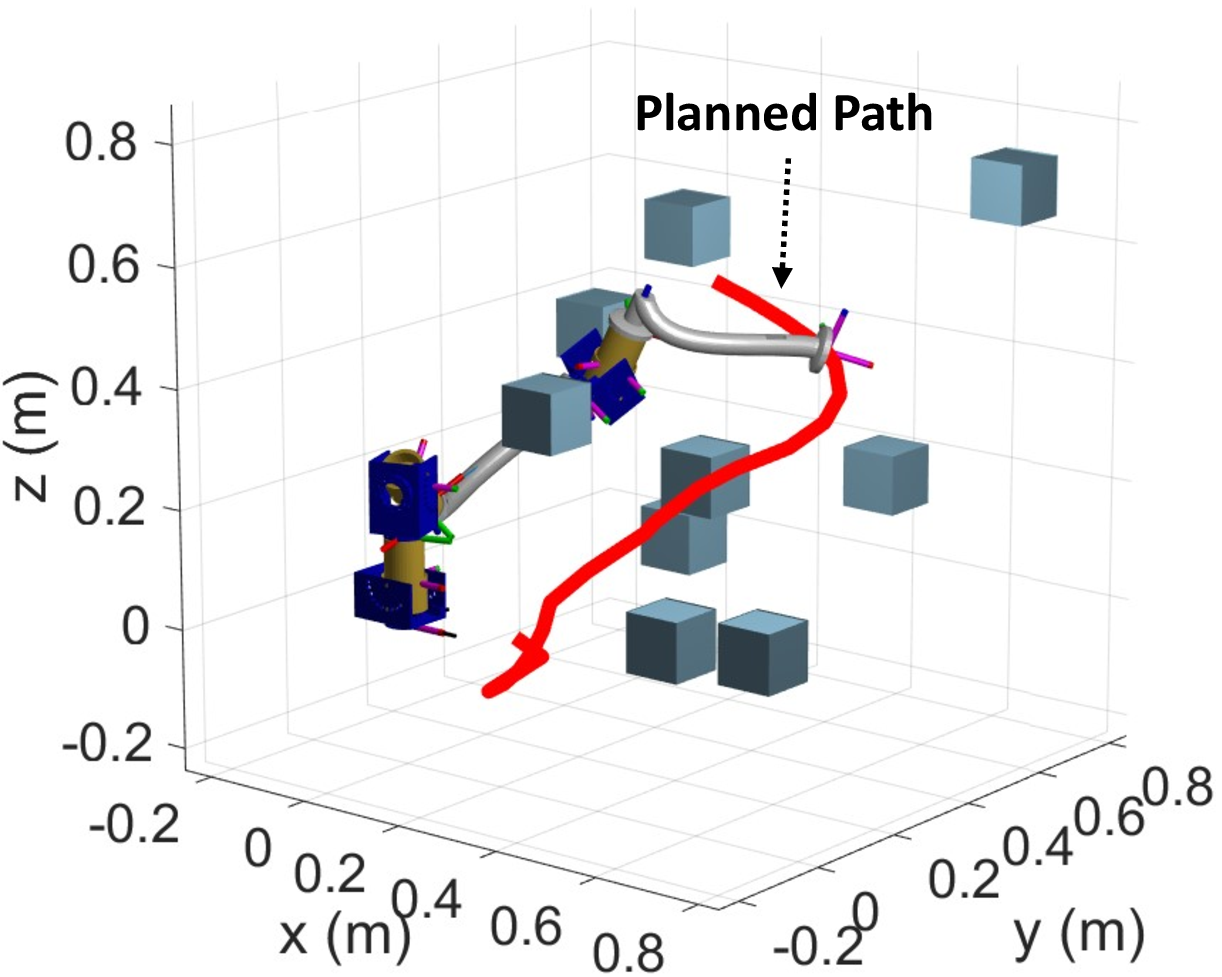}}
	~
	\subfigure[]{\includegraphics[width=2in]{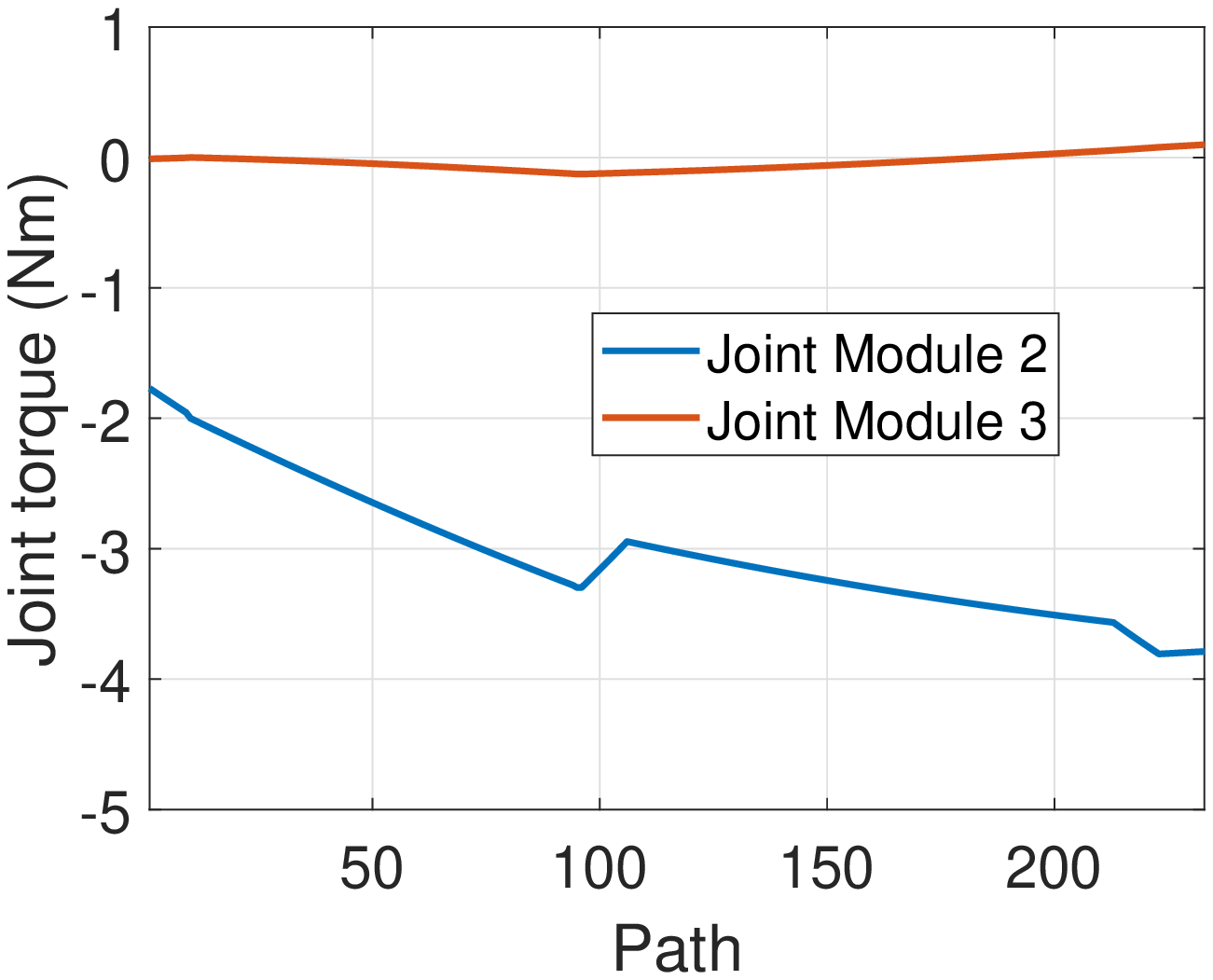}}
	
	\caption{(a) TSLs and the environment for the case study I-A. (b)$ H^1-H^4(-45^\circ)-H^3(45^\circ)$ modular composition synthesized for the given TSLs and environment. (c) Case I-A configuration reaching to TSLs in between the obstacles. (d) Planned path between the TSLs avoiding the obstacles of case I-A, (e) Computed joint torques over the path for joint module 2 ($H^4$) and joint module 3 ($H^3$) respectively.}
	\label{fig:case1A}
\end{figure}
\subsection{Case Study I-A}
A layout to represent a cluttered environment is made using closely spaced square boxes and the size of all the boxes is $0.1~m$, as shown in Fig.~\ref{fig:case1A}(a). At first, $4$ TSLs are selected where the manipulator has to reach with no requirement of the end-effector orientation at these locations. The coordinates of the TSLs with respect to the base location are $(0.1,0.6,0.5)$, $(0.4,0.6,0.4)$, $(0.6,-0.1,0.3)$ and $(0.4,-0.2,0.1)$ (in meters).

For the positional reachability in 3D space, a minimum of $3-$DoF configuration would be needed. The configuration reaching to these TSLs and with the minimum objective function comes out to be of $3-$DoF with sequence $H^1-H^4-H^3$ with $-45^\circ$ and $45^\circ$ of intersecting-twist angles in $H^4$ and $H^3$ respectively. Two links, $S_2$ and $C_2$ are used here with the sequence, as shown in Fig.~\ref{fig:case1A} (b). The configuration reaching to all the TSLs is shown in Fig.~\ref{fig:case1A}(c).

With the reachability for each TSL ensured, a complete path can be planned within the obstacles to move from one TSL to another. This is done by using \textit{RRT-connect} path planner for resulted modular composition in the prescribed collision environment made of boxes. The complete path, followed by the end-frame of the modular composition, between the four TSLs is shown in Fig.~\ref{fig:case1A}(d). The joint torques are computed over the path to check for the joint torque limits of the corresponding actuator, as shown in Fig.~\ref{fig:case1A}(e). The 3-DoF modular configuration in this case is composed of all H modules. As the nominal torque capacity of H module is $12~Nm$ (Table~\ref{tab:actuator_specs}), the current configuration is providing the satisfactory results as the maximum torque requirement is found nearly to be only $4~Nm$, as shown in Fig.~\ref{fig:case1A}(e). 

From the torque requirement data over the path, it can be noted that the requirement of the joint 3 is even less than the limit torque of L module, therefore, $H^3$ can be replaced by $L^3$ in this case. 
A standard 3-DoF configuration, even able to reach in this cluttered environment, as shown in Fig.~\ref{fig:collision_proof}, but is colliding with the obstacles. An observation worth to be noted through the proposed approach and through this case study is that, by introducing the unconventional (non-parallel and non-perpendicular twist) parameters in the design space, an unconventional configuration is generated which can reach to the given TSLs, and move between these TSLs while avoiding any collisions.

\begin{figure}[ht]
	\centering
	\subfigure[]{\includegraphics[width=2in]{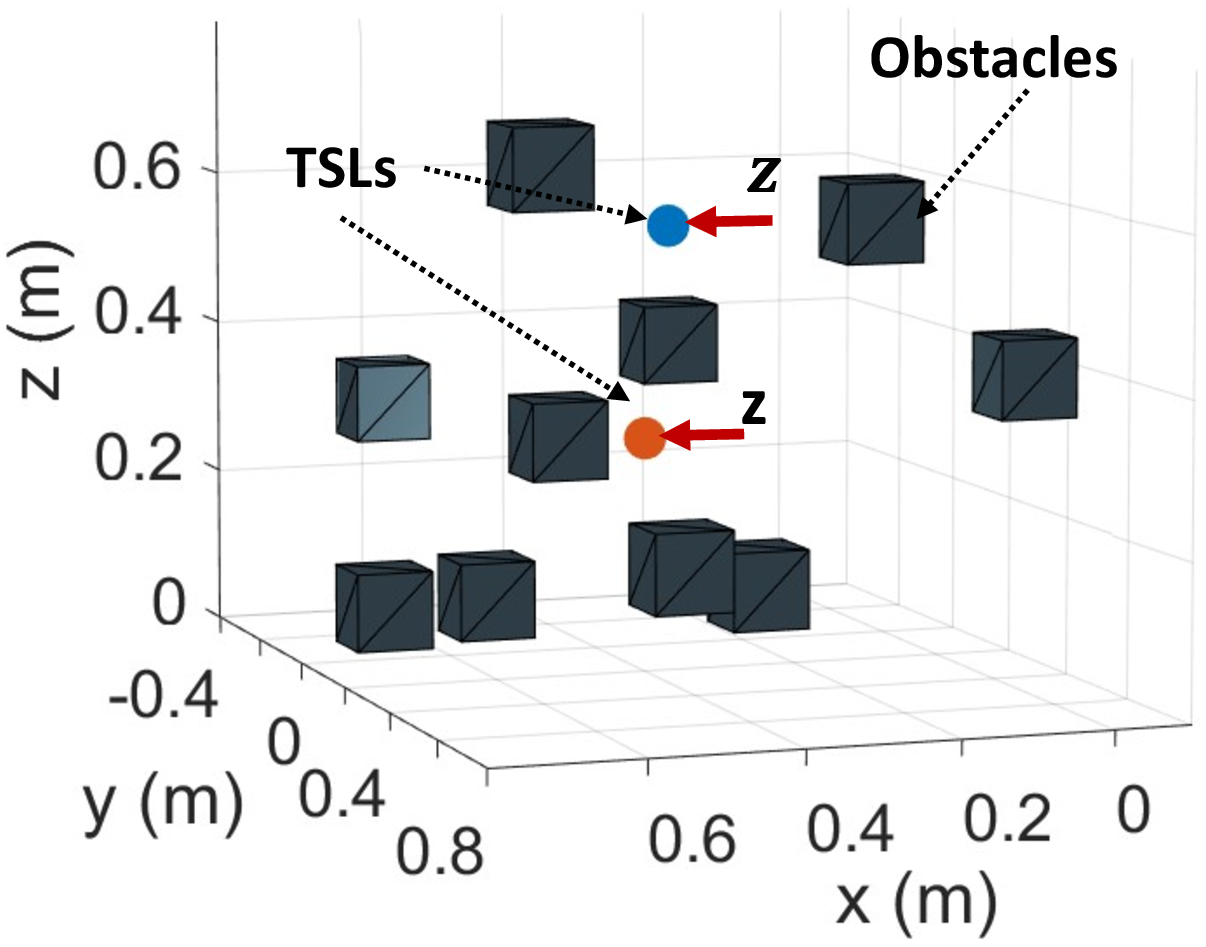}}
	~~~~~~~
	\subfigure[]{\includegraphics[width=1.35in]{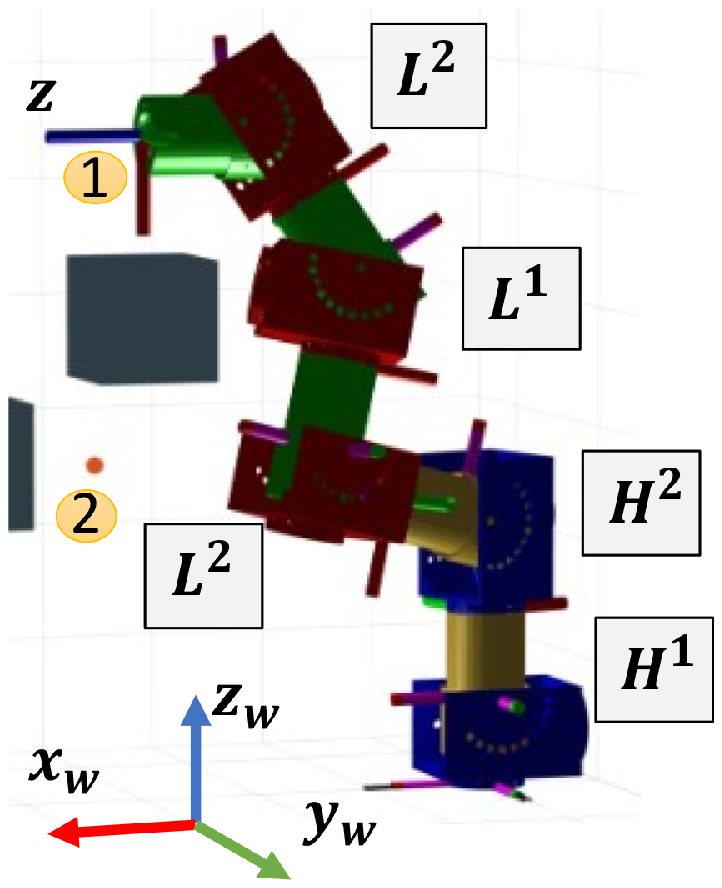}}	
	~
	\subfigure[]{\includegraphics[width=2.15in]{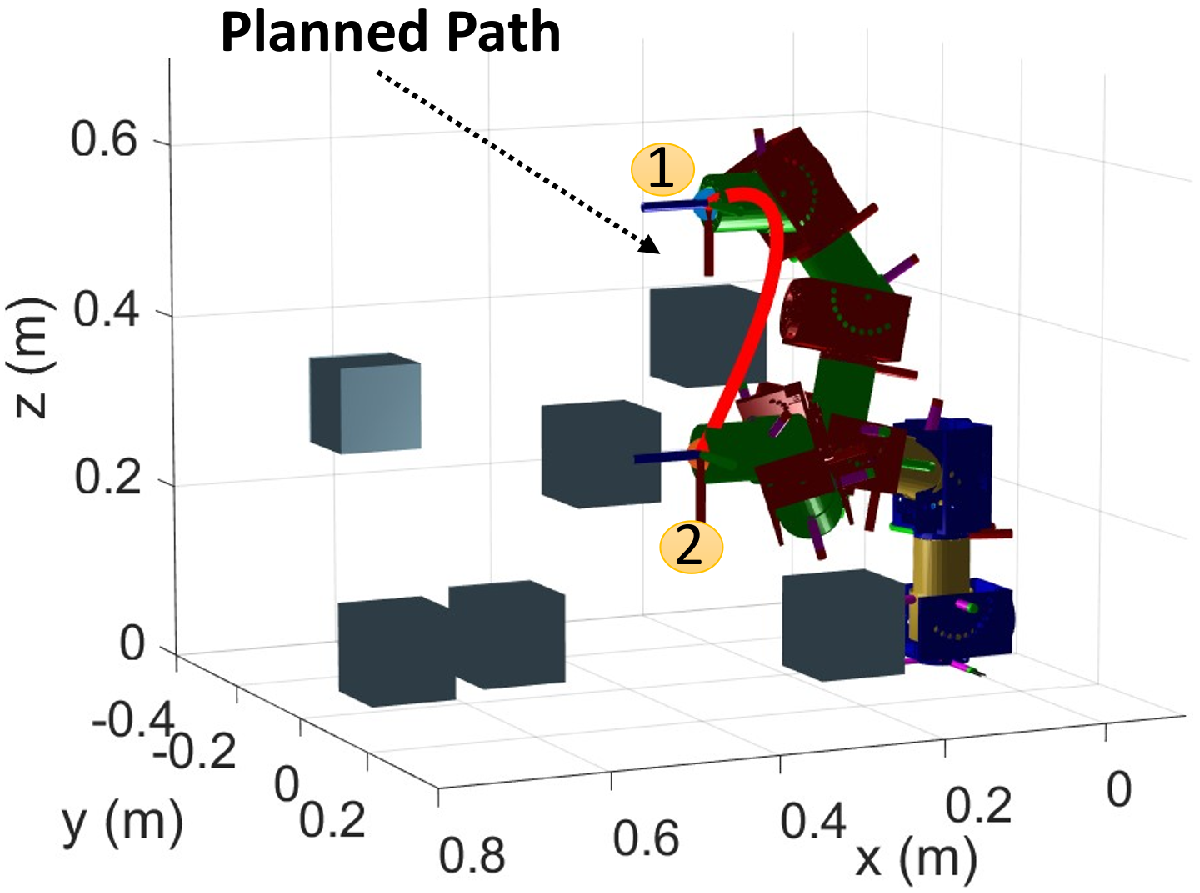}}
	~~~
	\subfigure[]{\includegraphics[width=1.85in]{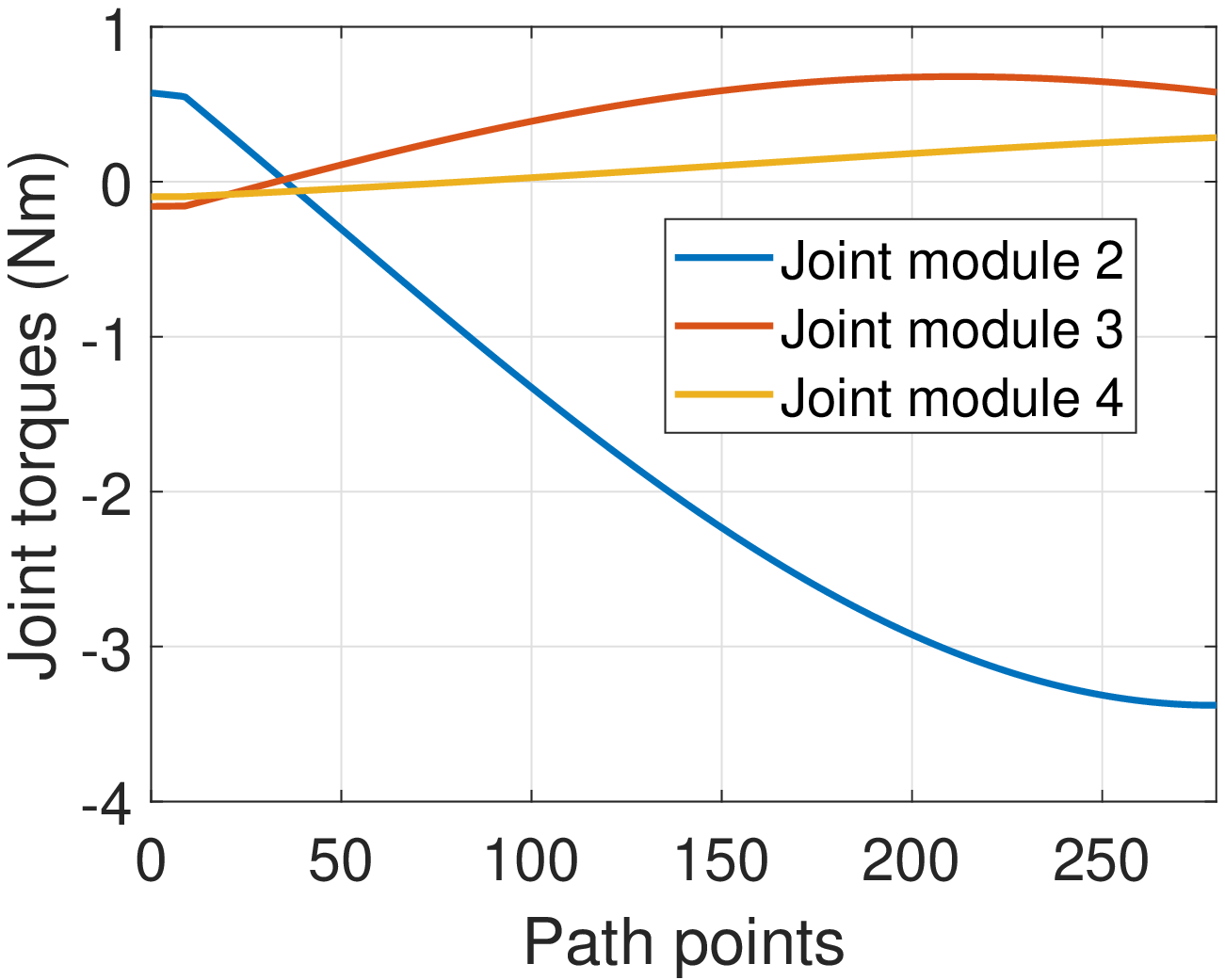}}
	
	\caption{(a) TSLs with orientation and the environment for the case study I-B. (b)$H^1-H^2(15^\circ)-L^2-L^1(-45^\circ)-L^2(60^\circ)$ modular composition synthesized for the given TSLs and environment. (c) Planned path for the modular Configuration from TSL \{1\} to TSL \{2\} for Case-IB. (d) Joint torque requirements for the joint module 2 ($H^1$), joint module 3 ($L^2$) and joint module 4 ($L^1$) over the planned path from \{1\} to \{2\}.}
	
	\label{fig:case1B}
\end{figure} 
\subsection{Case Study I-B}
In this case, the same layout of environment, as in Case study I-A, is considered but with 2 TSLs and with orientation constraints. 

The position of the TSLs are $(0.3,0,0.55)$ and $(0.4,0.2,0.3)$, and the orientations are fixed as the $z-$axis of the end-effector frame is facing towards a direction, as shown in Fig.~\ref{fig:case1B}(a), with roll, pitch and yaw angle for 3 TSLs as $(0,90^\circ,0)$ and $(0,90^\circ,0)$. In this case, a $5-$DoF modular composition is resulting as an optimal configuration with modular sequence of $H^1-H^2(15^\circ)-L^2-L^1(-45^\circ)-L^2(60^\circ)$, as shown in Fig.~\ref{fig:case1B}(b).% The zoomed view of the modular composition reaching to the one of the TSL and oriented in the given direction is shown in Fig.~\ref{fig:case1B}(c).
~To satisfy the positional and orientation requirements, 5-DoF configuration is generated  with no links. If the TSLs are selected farther, the link modules would be incorporated in the assembly, however not required in this case.

The post-processing involves generating the path within the cluttered environment to move from one TSL to another. Along with that, it is important to check if the joint torques are in the limits, with respect to actuator specifications, when the composition is following the generated path. The reachability of the modular composition to the two TSLs, and the planned path using \textit{RRT-connect} between the TSL \{1\} to TSL \{2\} is shown in Fig.~\ref{fig:case1B}(c). The torque requirements of the joint module 2, 3 and 4 are shown in Fig.~\ref{fig:case1B}(d). The joint torques of module 1 and 5 are negligible and thus are not shown in the plot. The 5-DoF modular configuration in this case is composed of all 2-H modules and 3-L modules. As the nominal torque capacity of H module is $12~Nm$ and of L module is $3.6~Nm$,  (Table~\ref{tab:actuator_specs}), the current configuration is providing the satisfactory results as the maximum torque requirement is found nearly to be only $3.5~Nm$, as shown in Fig.~\ref{fig:case1B}(d).
\begin{figure}[ht]
	\centering
	\subfigure[]{\includegraphics[width=1.25in]{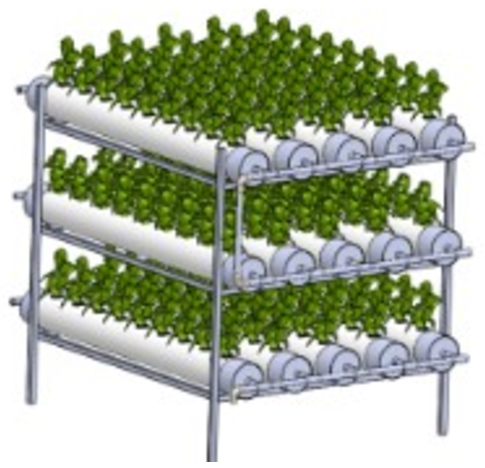}}
	~
	\subfigure[]{\includegraphics[width=1.9in]{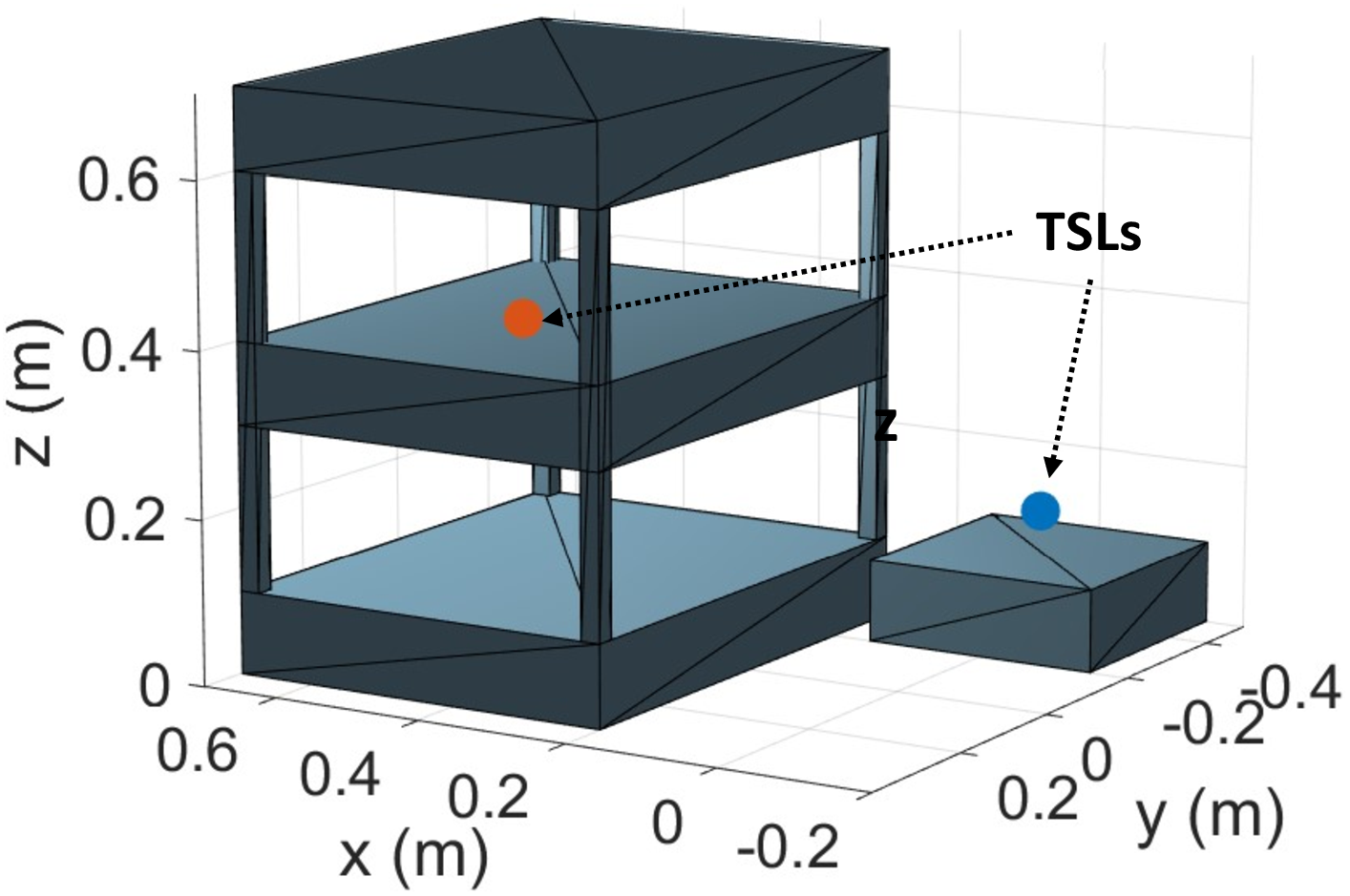}}
	~
	\subfigure[]{\includegraphics[width=0.8in]{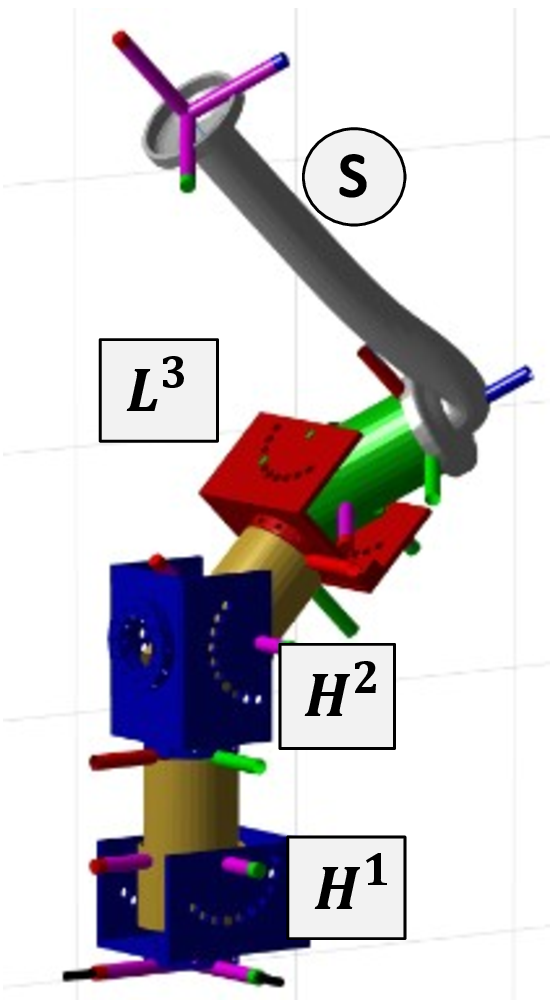}}
	~
	\subfigure[]{\includegraphics[width=2.15in]{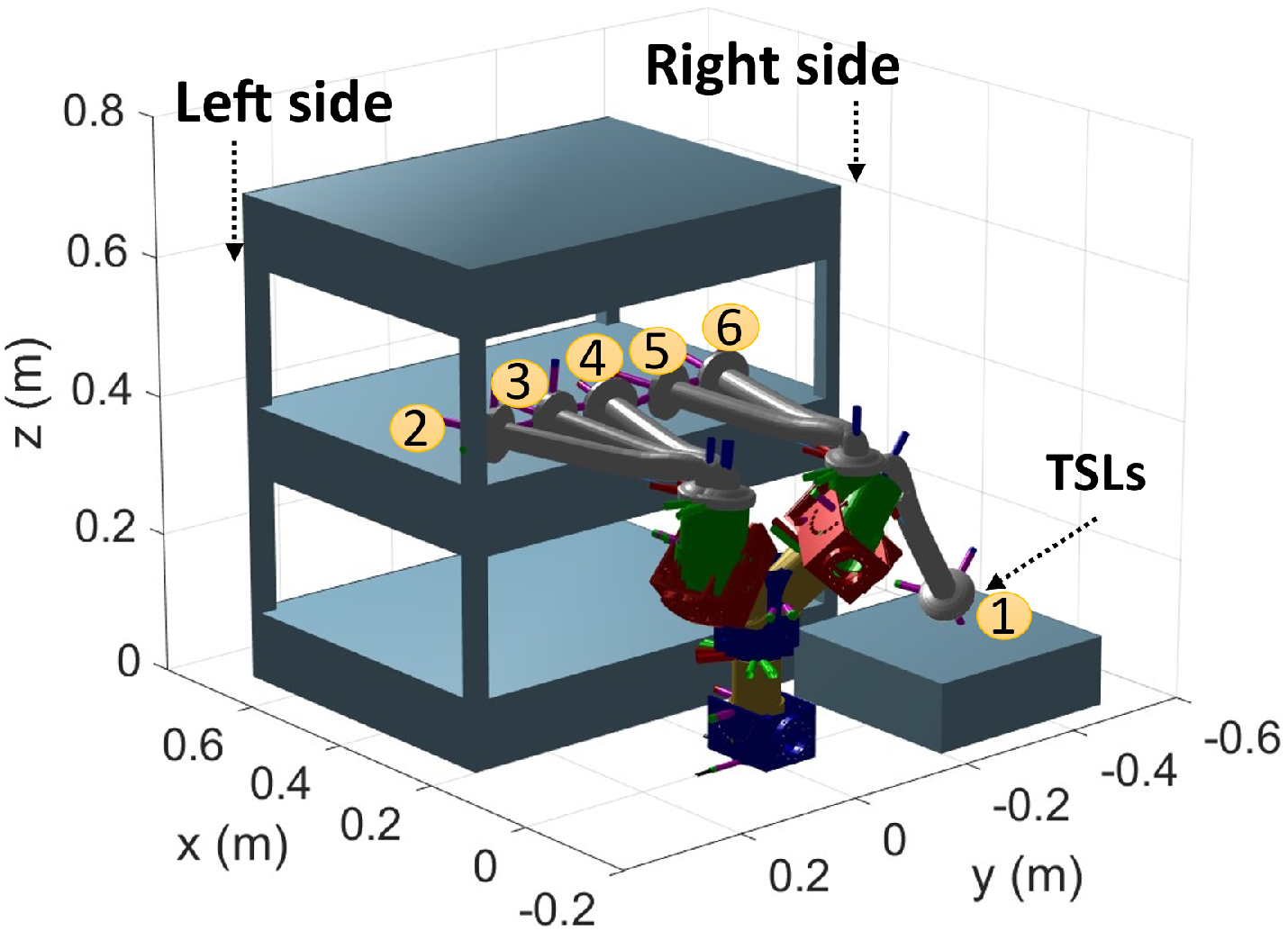}}
	~
	\subfigure[]{\includegraphics[width=2in]{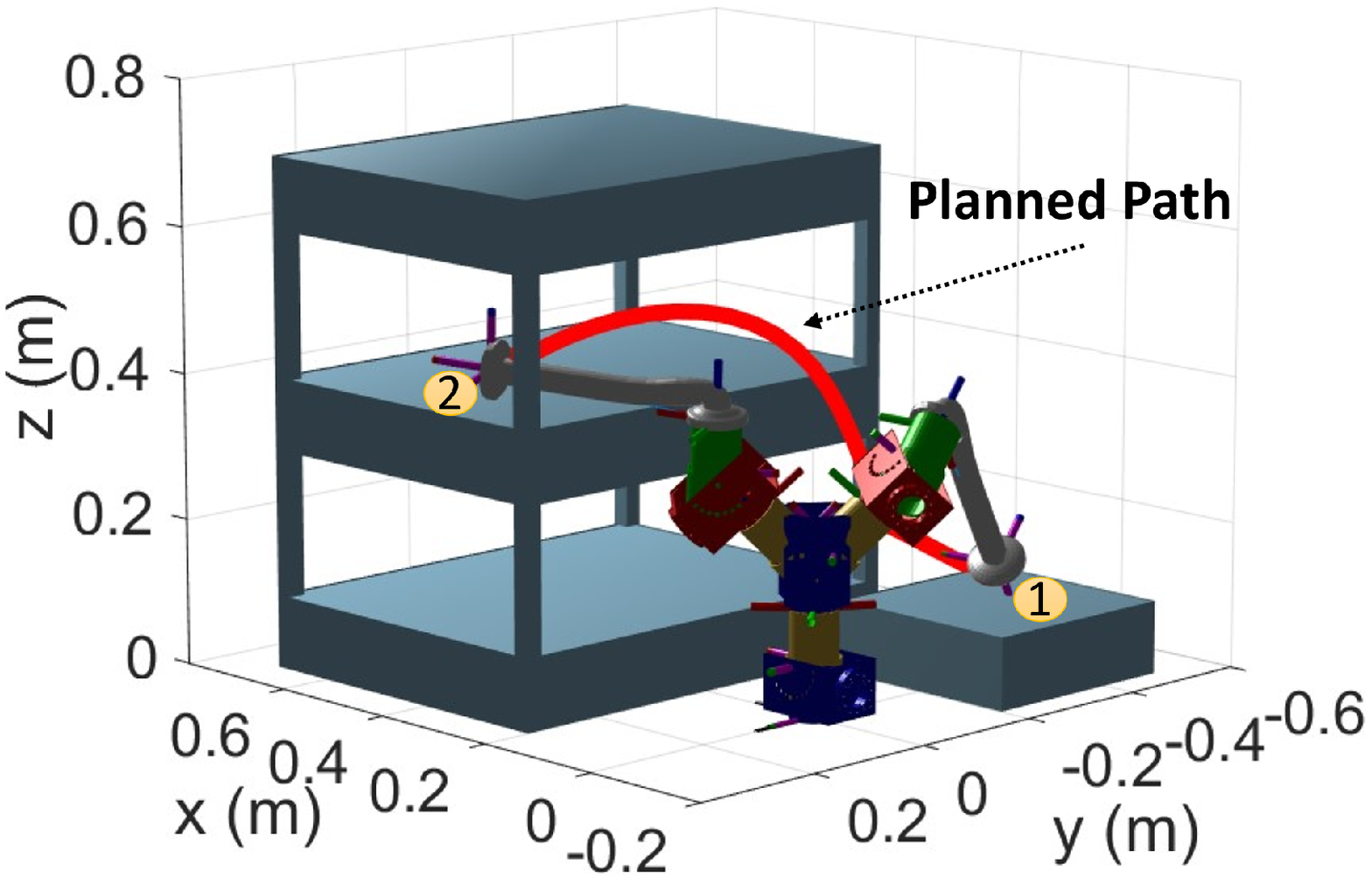}}
	~
	\subfigure[]{\includegraphics[width=1.95in]{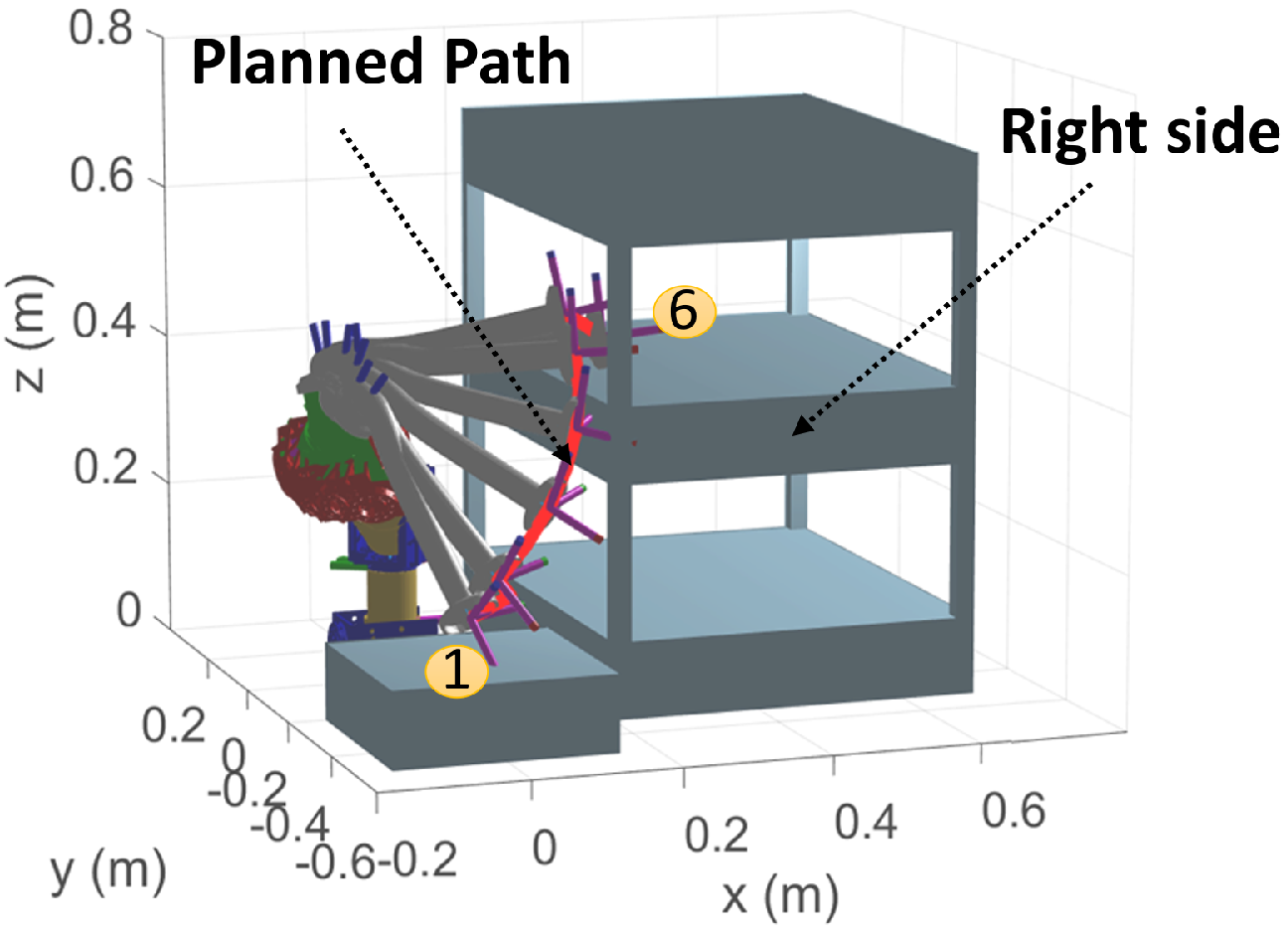}}
	~
	\subfigure[]{\includegraphics[width=1.6in]{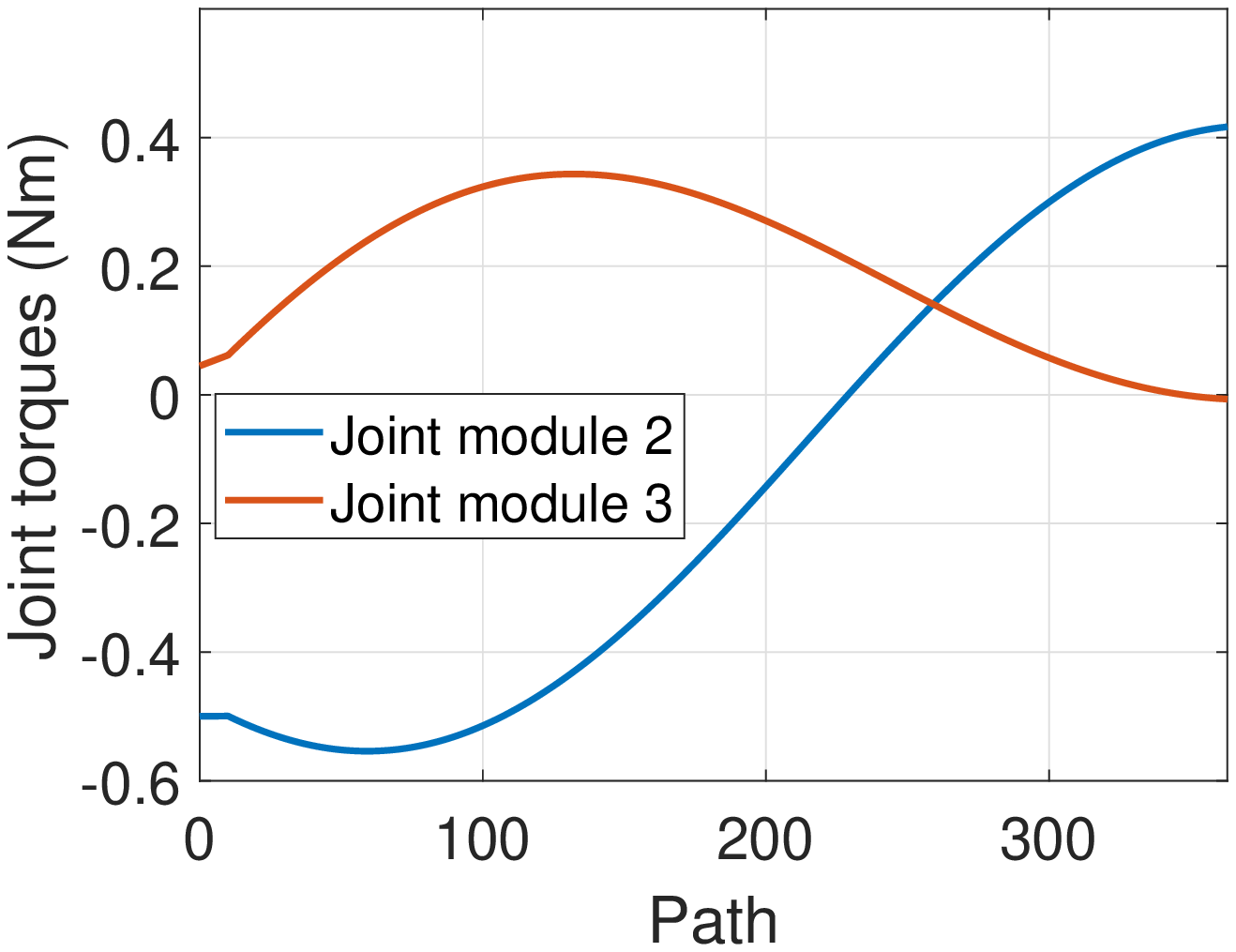}}
	
	\caption{(a) A realistic representation of a vertical farm. (b) Vertical farm setup with the TSLs to be reached. (c) 3-DoF configuraton with $H^1-H^2(45^\circ)-L^3(45^\circ)$. (d) Verification of the reachability of generated composition on the possible neighbouring TSLs. (e) Planned path for the modular Configuration from the seedling tray \{1\} to the transplanting location \{2\}. (f) Planned path for the modular Configuration from the seedling tray \{1\}to the transplanting location \{6\}. (g) Joint torque requirements for the joint module 2 ($H^2$)and joint module 3 ($L^3$) over the planned path.}
	\label{fig:csIIA}
\end{figure}  
\subsection{Case Study II-A}

In this case study, a realistic environment is considered for connecting the proposed strategy to the realistic applications having a specific layout. A replica of a cell of an agricultural vertical farm is generated of the size $0.55 \times 0.55 \times 0.7~m$ with 3 shelves for plantation, as shown in Fig.~\ref{fig:csIIA}(a) and (b). A seedling tray is generated near the base of the robot of size $0.3\times0.3\times0.1~m$, from where the robot has to start with the transplanting. Two of the TSLs are selected here as per the requirement of robotic assistance in agricultural vertical farms.

First is the pick-up location of the seedlings from the seedling tray, kept besides the manipulator. Second is the drop location of the seedlings into the pots of the vertical farm. 
For the pick and place task in this case, a 3-DoF configuration with one $S_1$ link is evolved with sequence as, $H^1-H^2-L^3$ having $45^\circ$ of intersecting-twist angles in joint 2 and 3, as shown in Fig.~\ref{fig:csIIA}(b). The planned motion from pick-up location to the drop location is shown in Fig.~\ref{fig:csIIA}(c). 

As the transplanting process involves picking and placing the seedlings from the tray to the vertical farm, similar TSLs can be considered along the length of the setup. To demonstrate this, verification of the reachability for the modular composition over the neighbouring TSLs (\{2\} to \{6\}) is done, as shown in Fig.~\ref{fig:csIIA}(d). The same composition is able to reach all the TSLs. This shows that, it is also possible to execute the program for lesser number of TSLs, as it saves the time of computation, and then validating the solution for all the required TSLs through post-processing.

The planned motion using \textit{RRT-connect} planner within the vertical farm environment from the pick-up location \{1\} to the drop location \{2\}, is shown in Fig.~\ref{fig:csIIA}(e), and from pick-up location \{1\} to the drop location \{6\} is shown in Fig.~\ref{fig:csIIA}(f). The joint torques are computed over the path, from \{1\} to \{2\}, to check for the joint torque limits of the corresponding actuator, as shown in Fig.~\ref{fig:csIIA}(g). The 3-DoF modular configuration in this case is composed of all 2-H modules and 1-L module and have the satisfactory requirements of joint torque as the maximum torque requirement is found nearly to be only $0.6~Nm$, as shown in Fig.~\ref{fig:csIIA}(g).
\subsection{Case Study II-B}
\begin{figure}[h!]
	\centering
	\subfigure[]{\includegraphics[width=2in]{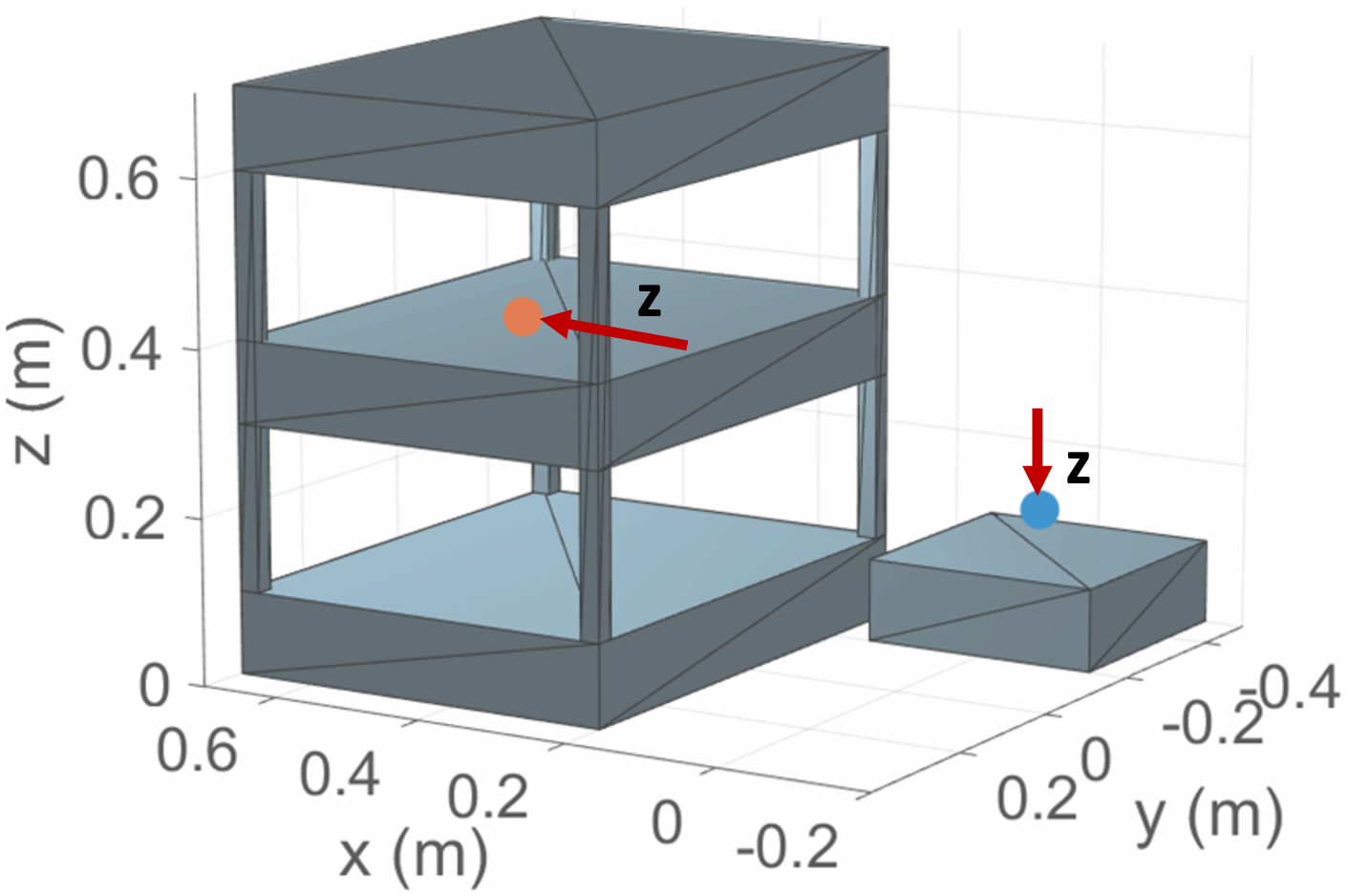}}
	~
	\subfigure[]{\includegraphics[width=2in]{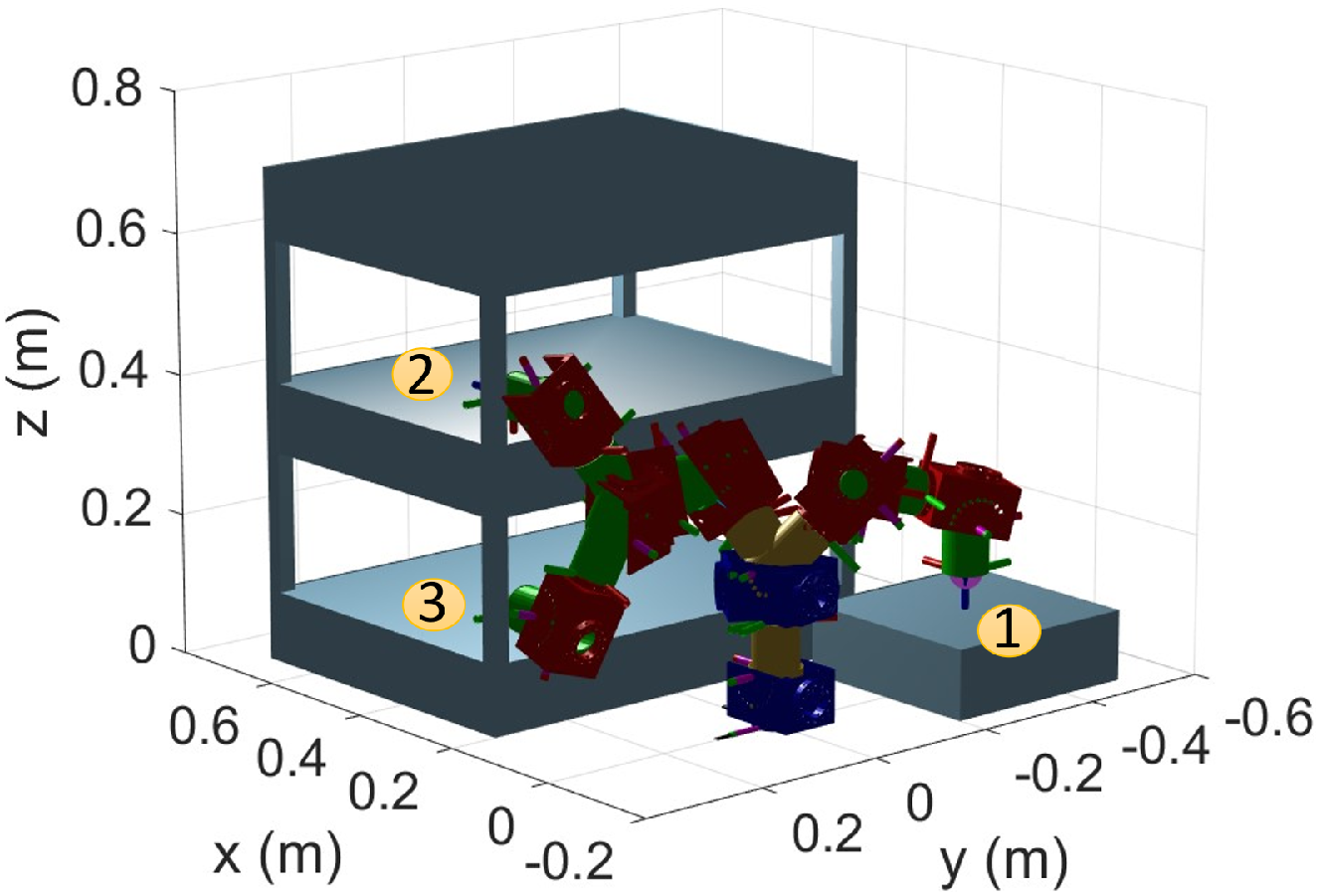}}
	~
	\subfigure[]{\includegraphics[width=2in]{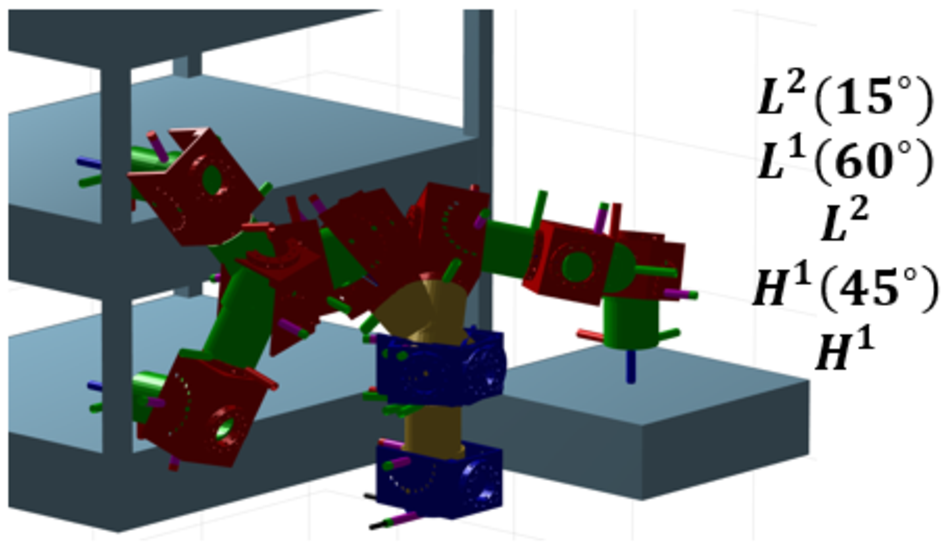}}
	~
	\subfigure[]{\includegraphics[width=2in]{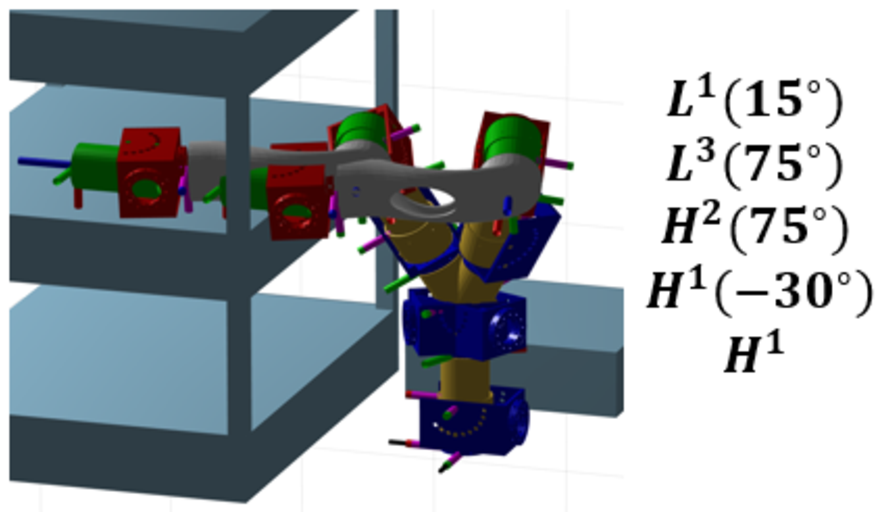}}
	~
	\subfigure[]{\includegraphics[width=2.25in]{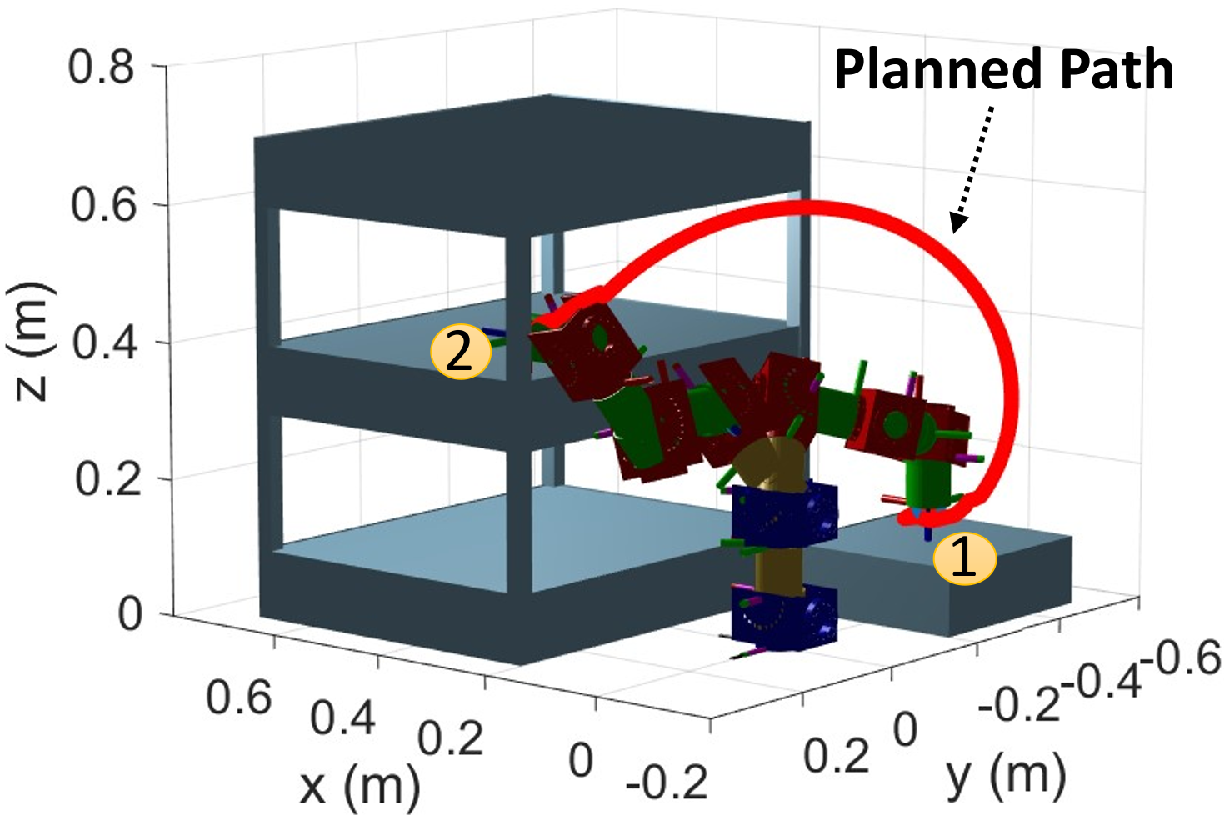}}
	~
	\subfigure[]{\includegraphics[width=1.75in]{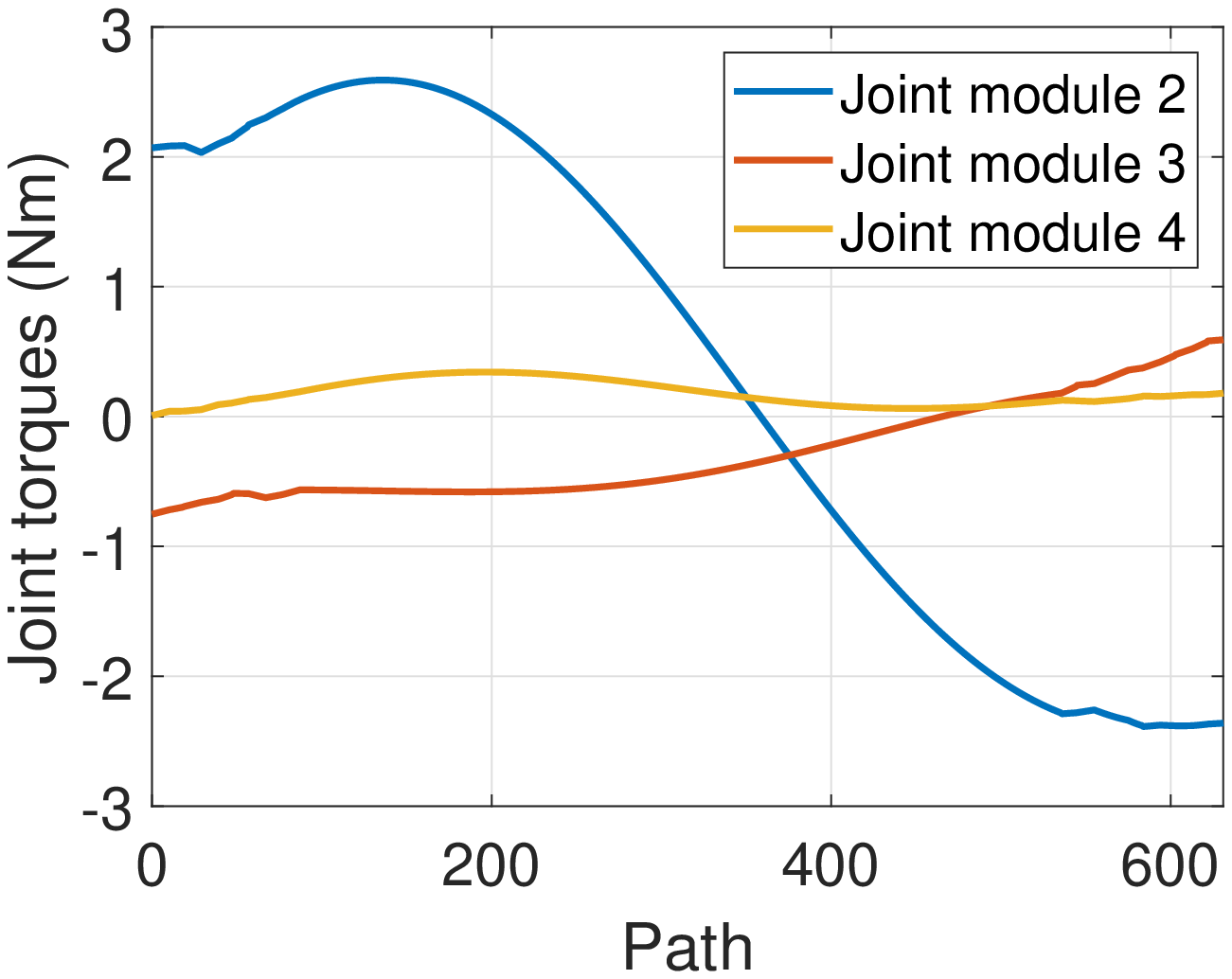}}
	~
	\subfigure[]{\includegraphics[width=2.25in]{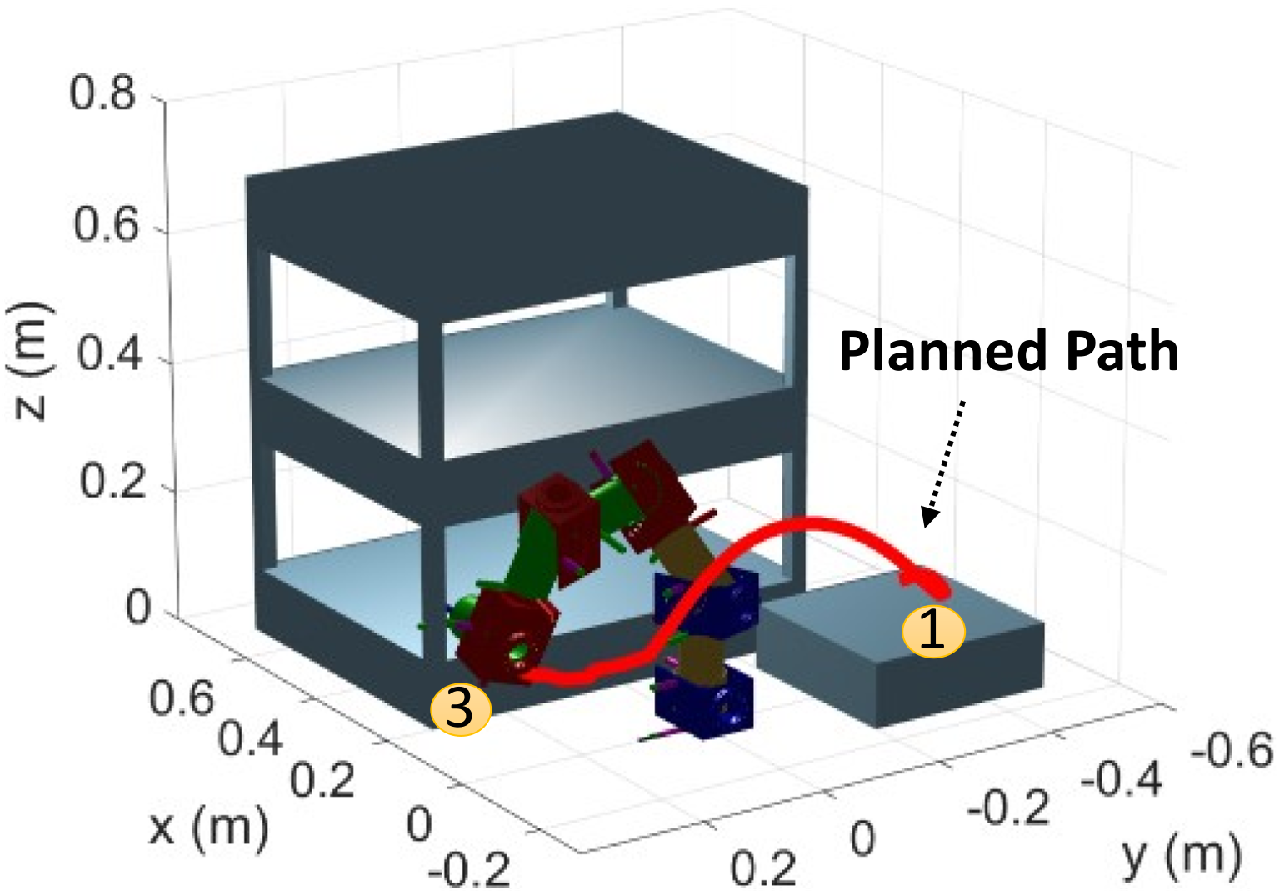}}
	~
	\subfigure[]{\includegraphics[width=1.75in]{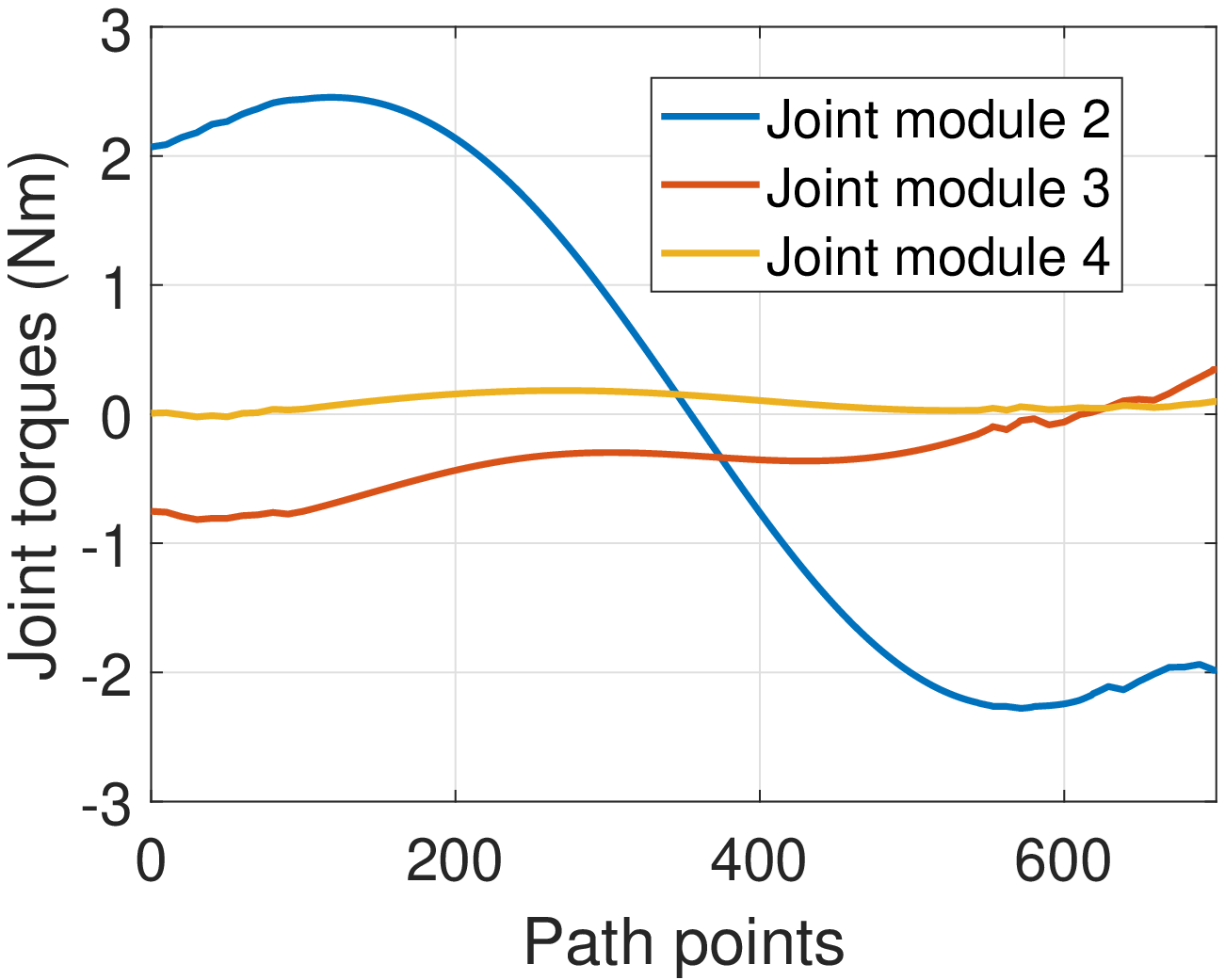}}
	\caption{(a) TSLs in the vertical farm with orientational constraints. (b)  $5-$DoF modular composition reaching and oriented to all 3 TSLs as prescribed. (c) $5-$DoF modular composition with $H^1-H^1(45^\circ)-L^2-L^1(60^\circ)-L^2(15^\circ)$ able to reach all the TSLs, (d) $5-$DoF modular composition with $H^1-H^1(-30^\circ)-H^2(75^\circ)-L^3(75^\circ)-L^1(15^\circ)$ with link module reaching inside the vertical farm. (e) Planned path for the modular Configuration from TSL \{1\} to TSL \{2\}. (f) Joint torque requirements for the joint module 2 ($H^1$), joint module 3 ($L^2$) and joint module 4 ($L^1$) over the planned path from \{1\} to \{2\}. (g) Planned path for the modular Configuration from TSL \{1\} to TSL \{3\}. (h) Joint torque requirements for the joint module 2 ($H^1$), joint module 3 ($L^2$) and joint module 4 ($L^1$) over the planned path from \{1\} to \{3\}.}
	\label{fig:csIIb}
\end{figure}
In this case, the same layout of the vertical farm is considered. The orientation constraints are added with the TSLs as the $z-$axis of the end-effector of the modular composition is to be aligned with $z-$axes of the TSLs, as shown in Fig.~\ref{fig:csIIb}(a).

With the orientational constraints added with the positional constraints, the solution came out to be of $5-$DoF modular composition with no links. The configuration is composed of $H^1-H^1(45^\circ)-L^2-L^1(60^\circ)-L^2(15^\circ)$,  as shown in Fig.~\ref{fig:csIIb}(b) and (c). One more possibility of task is to reach far and inside the vertical farm for which the optimal synthesis proposed $5-$DoF configuration with a link, as $H^1-H^1(-30^\circ)-H^2(75^\circ)-L^3(75^\circ)-L^1(15^\circ)$, as shown in Fig.~\ref{fig:csIIb}(d).

The planned motion using \textit{RRT-connect} planner with in the vertical farm environment from TSL \{1\} to TSL \{2\}, is shown in Fig.~\ref{fig:csIIb}(e), and from TSL \{1\} to TSL \{3\} is shown in Fig.~\ref{fig:csIIb}(g). The joint torques are computed over the path, from \{1\} to \{2\} and from \{1\} to \{3\}, to check for the joint torque limits of the corresponding actuator, as shown in Fig.~\ref{fig:csIIb}(f) and Fig.~\ref{fig:csIIb}(h). The 5-DoF modular configuration in this case is composed of all 2-H modules and 3-L modules. As the nominal torque capacity of H module is $12~Nm$ and of L module is $3.6~Nm$,  (Table~\ref{tab:actuator_specs}), the current configuration is providing the satisfactory results as the maximum torque requirement is found nearly to be only $2.5~Nm$, as shown in Fig.~\ref{fig:csIIb}(f) and Fig.~\ref{fig:csIIb}(h).

\subsection{Discussions}
\begin{figure}[]
	\centering
	\subfigure[]{\includegraphics[width=0.85\columnwidth]{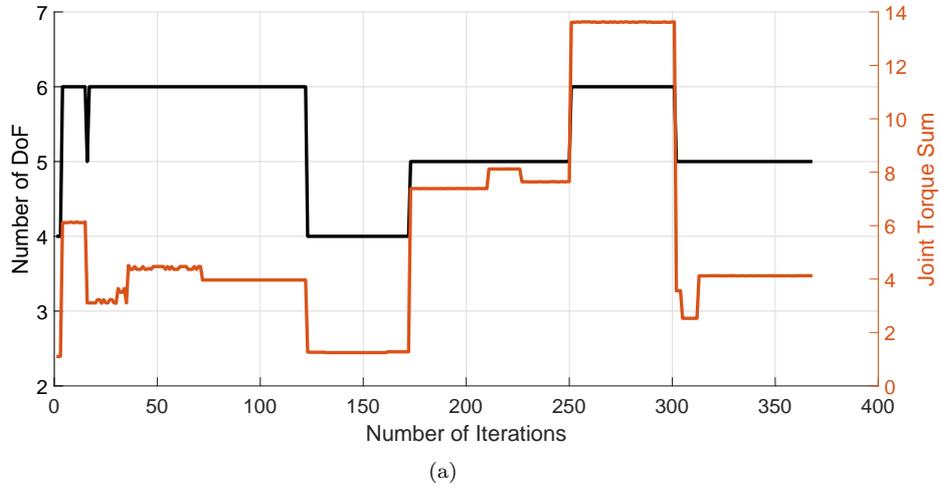}}
	\\
	\subfigure[]{\includegraphics[width=2.8in]{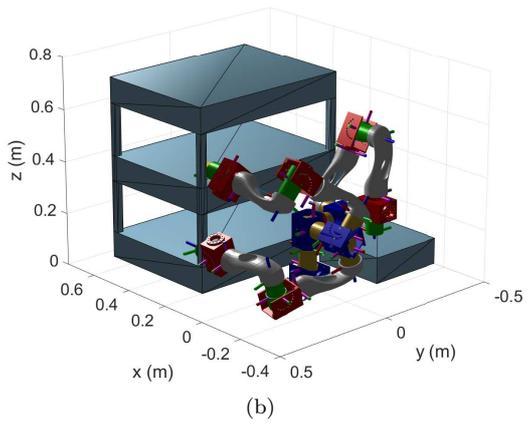}}
	~
	\subfigure[]{\includegraphics[width=2.6in]{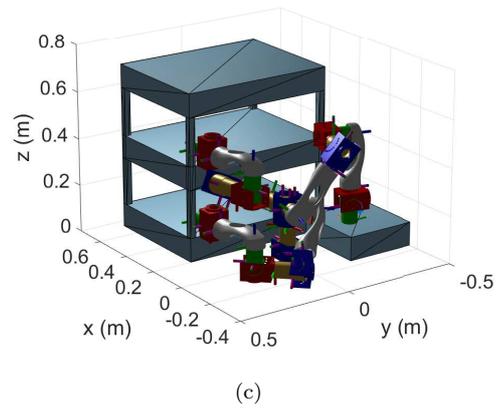}}
	\\
	\subfigure[]{\includegraphics[width=2.5in]{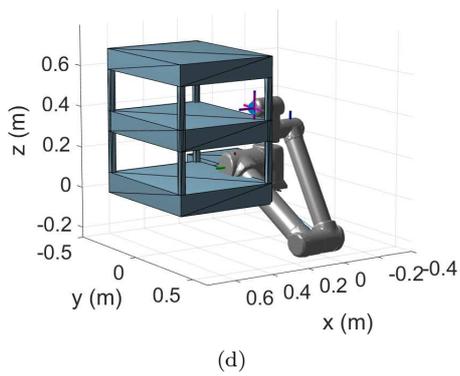}}
	~
	\subfigure[]{\includegraphics[width=2.4in]{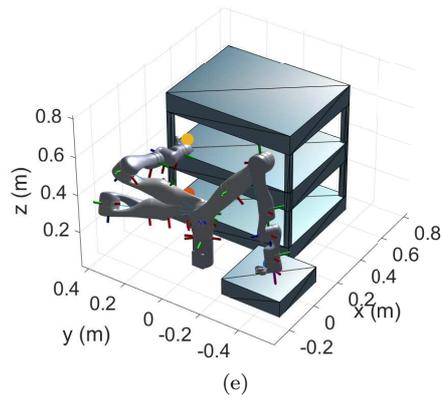}}
	
	\caption{(a) Variation of DoF and the joint torque sum of generated modular compositions during an optimization routine of case II-B, (b) 5-DoF conventional modular composition with sequence $H^1-H^2-H^4-L^4-L^1$ generated for TSLs and workspace as in case II-B, (c) 5-DoF conventional modular composition with sequence $H^1-H^4-H^2-L^4-L^1$ generated for comparison in terms of joint torque sum, (d) \textit{UR10} configuration in the given workcell as in case II-B, (e) \textit{Kinova-gen3} composition in the given workcell.}
	\label{fig:csIII}
\end{figure}

The optimization problems in this work are solved using a single objective function, however, one can use multi-objectives for the required tasks, such as reachability and joint torques, manipulator performance measures as the multiple objectives. The genetic algorithm is a heuristic technique containing an element of randomness, therefore, it is possible that solutions reached are only sub-optimal but well performing for the specific execution of optimization algorithm. In future, further investigations towards global optimality are planned. The optimization problem formulated in this work is novel in terms of having the variables in DoF, unconventional twist angles, types of modular units and links. Figure~\ref{fig:csIII}(a) shows the variation of the DoF and the Joint torque sum of the modular compositions during the optimization routine for the case study II-B. \\
To show the applicability of the proposed algorithm even for the conventional modules, case II-B is evaluated for conventional configurations, i.e., only parallel and perpendicular jointed configurations. To perform this, the design variables of the twist angles have been discarded and only the 4 modular units ($H^1, ~H^2, ~H^3, ~H^4$), similarly for L variant, considering the assembly rules are used in the optimization routines. Four cases are shown in Figs.~\ref{fig:csIII}(b)-(e), to illustrate the approach. It has been noticed that, the compositions generated using the proposed modules are of 5-DoF with parallel or perpendicular axes (Conventional/Traditional manipulator design) and the modular sequence comes out to be $H^1-H^2-H^4-L^4-L^1$ and $H^1-H^4-H^2-L^4-L^1$ respectively as shown in Figs.~\ref{fig:csIII}(b) and (c). It is worth to note that, even though both are 5-DoF compositions with the second and the third modular unit swapped with each other, the joint torque sum of the case III-A (6.0760) is less then the case III-B (7.55). The torque sum of unconventional configuration generated originally for the same TSLs and the workspace in case II-B is 4.12, as shown in Fig.~\ref{fig:csIII}(a), which is less then both the conventional configurations. The 5-DoF conventional configurations in both the cases are also having an error of atleast 0.05 m while reaching the TSLs. To further compare the results with the traditional manipulators, two 6-DoF conventional configurations have been considered to work upon the same TSLs and in the same environment as in case II-B. It has been noticed that both the manipulator configurations have failed to accomplish the task as shown through Figs.~\ref{fig:csIII}(d) and (e). The \textit{UR10} configuration in Fig.~\ref{fig:csIII}(d) is able to reach the TSLs but is colliding with the given environment. The \textit{Kinova-gen3} configuration is not even able to reach the TSLs. This highlights the novelty and the contribution of the proposed approach in which domain of the design parameters is further expanded through the twist angles. The generated unconventional modular compositions are proven to be useful in such cluttered workspaces as shown through the case studies in section~\ref{sec:results}.

\begin{figure}[t]
	\centering
	\subfigure[]{\includegraphics[width=1.85in]{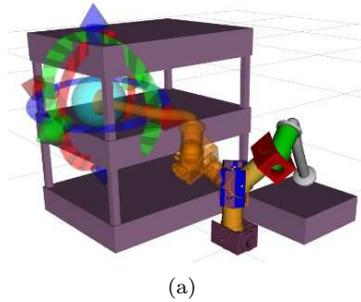}} 
	\caption{Demonstration of utilizing the optimal configuration in ROS for motion control}
	\label{fig:ros_implement}
\end{figure}

The optimal configuration i.e., the generated URDF file through the optimization process, is ready to be used in the Robotic Operating System (ROS) platform for the control of developed manipulator, as shown in Fig~\ref{fig:ros_implement}. The corresponding files for the configuration of the controller and actuators are to be developed along with the URDF in ROS, the details can be found in~\cite{dograUnified2022}.

\section{Conclusion}\label{sec:conclusion}
This paper proposed an approach for the optimal generation of custom configurations required to accomplish different set of tasks in different cluttered environments. The optimization problem has been developed to search for an optimal composition of the modular elements from a given modular library along with the robotic parameters required for configuration realization. The modular library is specifically designed to adapt the unconventional parameters which are not much used in the literature such as twist angles, and the curved links. The problem is solved using genetic algorithm with design variables as discrete variables from the modular library itself. It is shown through the results that, selection of the twist parameters in the design variable expands the range of the design space, which are aiding in finding the solutions for cluttered environment. The results are coming as unified models of the modular compositions which can be used directly in the ROS platform for the motion planning and control of the assembled modular configurations. Therefore, the approach gives an end-to-end solution for realizing the custom configurations from synthesis to deployment in a real-scenario.

\bibliography{ms}

\begin{thebibliography}{10}
\expandafter\ifx\csname url\endcsname\relax
  \def\url#1{\texttt{#1}}\fi
\expandafter\ifx\csname urlprefix\endcsname\relax\def\urlprefix{URL }\fi
\expandafter\ifx\csname href\endcsname\relax
  \def\href#1#2{#2} \def\path#1{#1}\fi

\bibitem{bi2020framework}
Z.~Bi, Z.~Miao, B.~Zhang, C.~W. Zhang, Framework for performance assessment of
  heterogeneous robotic systems, IEEE Systems Journal 15~(1) (2020) 1191--1201.
\newblock \href {https://doi.org/10.1109/jsyst.2020.2990892}
  {\path{doi:10.1109/jsyst.2020.2990892}}.

\bibitem{singh2018modular}
S.~Singh, A.~Singla, E.~Singla, Modular manipulators for cluttered
  environments: A task-based configuration design approach, Journal of
  Mechanisms and Robotics 10~(5) (2018) 051010.
\newblock \href {https://doi.org/10.1115/1.4040633}
  {\path{doi:10.1115/1.4040633}}.

\bibitem{althoff2019effortless}
M.~Althoff, A.~Giusti, S.~B. Liu, A.~Pereira, Effortless creation of safe
  robots from modules through self-programming and self-verification, Science
  Robotics 4~(31) (2019).
\newblock \href {https://doi.org/10.1126/scirobotics.aaw1924}
  {\path{doi:10.1126/scirobotics.aaw1924}}.

\bibitem{Modman2020}
A.~Yun, D.~Moon, J.~Ha, S.~Kang, W.~Lee, Modman: An advanced reconfigurable
  manipulator system with genderless connector and automatic kinematic modeling
  algorithm, IEEE Robotics and Automation Letters 5~(3) (2020) 4225--4232.
\newblock \href {https://doi.org/10.1109/lra.2020.2994486}
  {\path{doi:10.1109/lra.2020.2994486}}.

\bibitem{hong2017joint}
S.~Hong, C.~Cho, H.~Lee, S.~Kang, W.~Lee, Joint configuration for physically
  safe human--robot interaction of serial-chain manipulators, Mechanism and
  Machine Theory 107 (2017) 246--260.
\newblock \href {https://doi.org/10.1016/j.mechmachtheory.2016.10.002}
  {\path{doi:10.1016/j.mechmachtheory.2016.10.002}}.

\bibitem{valsamos2014kinematic}
C.~Valsamos, V.~Moulianitis, N.~Aspragathos, Kinematic synthesis of structures
  for metamorphic serial manipulators, Journal of Mechanisms and Robotics 6~(4)
  (2014) 041005.
\newblock \href {https://doi.org/10.1115/1.4027741}
  {\path{doi:10.1115/1.4027741}}.

\bibitem{acaccia2008modular}
G.~Acaccia, L.~Bruzzone, R.~Razzoli, A modular robotic system for industrial
  applications, Assembly Automation 28~(2) (2008) 151--162.
\newblock \href {https://doi.org/10.1108/01445150810863734}
  {\path{doi:10.1108/01445150810863734}}.

\bibitem{chen1997kinematic}
I.-M. Chen, G.~Yang, Kinematic calibration of modular reconfigurable robots
  using product-of-exponentials formula, Journal of robotic systems 14~(11)
  (1997) 807--821.
\newblock \href
  {https://doi.org/10.1002/(sici)1097-4563(199711)14:11<807::aid-rob4>3.0.co;2-y}
  {\path{doi:10.1002/(sici)1097-4563(199711)14:11<807::aid-rob4>3.0.co;2-y}}.

\bibitem{brandstotter2018task}
M.~Brandst{\"o}tter, P.~Gallina, S.~Seriani, M.~Hofbaur, Task-dependent
  structural modifications on reconfigurable general serial manipulators, in:
  International Conference on Robotics in Alpe-Adria Danube Region, Springer,
  2018, pp. 316--324.
\newblock \href {https://doi.org/10.1007/978-3-030-00232-9_33}
  {\path{doi:10.1007/978-3-030-00232-9_33}}.

\bibitem{stravopodis2020rectilinear}
N.~Stravopodis, V.~Moulianitis, Rectilinear tasks optimization of a modular
  serial metamorphic manipulator, Journal of Mechanisms and Robotics 13~(1)
  (2020).
\newblock \href {https://doi.org/10.1115/1.4047727}
  {\path{doi:10.1115/1.4047727}}.

\bibitem{dograJMD2021}
A.~Dogra, S.~Sekhar~Padhee, E.~Singla, An optimal architectural design for
  unconventional modular reconfigurable manipulation system, Journal of
  Mechanical Design 143~(6) (2021).
\newblock \href {https://doi.org/10.1115/1.4048821}
  {\path{doi:10.1115/1.4048821}}.

\bibitem{patel2015task}
S.~Patel, T.~Sobh, Task based synthesis of serial manipulators, Journal of
  advanced research 6~(3) (2015) 479--492.
\newblock \href {https://doi.org/10.1016/j.jare.2014.12.006}
  {\path{doi:10.1016/j.jare.2014.12.006}}.

\bibitem{tabandeh2016memetic}
S.~Tabandeh, W.~Melek, M.~Biglarbegian, S.-h.~P. Won, C.~Clark, A memetic
  algorithm approach for solving the task-based configuration optimization
  problem in serial modular and reconfigurable robots, Robotica 34~(9) (2016)
  1979--2008.
\newblock \href {https://doi.org/10.1017/S0263574714002690}
  {\path{doi:10.1017/S0263574714002690}}.

\bibitem{whitman2018task}
J.~Whitman, H.~Choset, Task-specific manipulator design and trajectory
  synthesis, IEEE Robotics and Automation Letters 4~(2) (2018) 301--308.
\newblock \href {https://doi.org/10.1109/lra.2018.2890206}
  {\path{doi:10.1109/lra.2018.2890206}}.

\bibitem{campos2019task}
T.~Campos, J.~P. Inala, A.~Solar-Lezama, H.~Kress-Gazit, Task-based design of
  ad-hoc modular manipulators, in: 2019 International Conference on Robotics
  and Automation (ICRA), IEEE, 2019, pp. 6058--6064.
\newblock \href {https://doi.org/10.1109/icra.2019.8794171}
  {\path{doi:10.1109/icra.2019.8794171}}.

\bibitem{chen1998enumerating}
I.-M. Chen, J.~W. Burdick, Enumerating the non-isomorphic assembly
  configurations of modular robotic systems, The International Journal of
  Robotics Research 17~(7) (1998) 702--719.
\newblock \href {https://doi.org/10.1177/027836499801700702}
  {\path{doi:10.1177/027836499801700702}}.

\bibitem{bi2001concurrent}
Z.~Bi, W.-J. Zhang, Concurrent optimal design of modular robotic configuration,
  Journal of Robotic systems 18~(2) (2001) 77--87.
\newblock \href
  {https://doi.org/10.1002/1097-4563(200102)18:2<77::aid-rob1007>3.0.co;2-a}
  {\path{doi:10.1002/1097-4563(200102)18:2<77::aid-rob1007>3.0.co;2-a}}.

\bibitem{chung1997task}
W.~K. Chung, J.~Han, Y.~Youm, S.~Kim, Task based design of modular robot
  manipulator using efficient genetic algorithm, in: Proceedings of
  International Conference on Robotics and Automation, Vol.~1, IEEE, 1997, pp.
  507--512.
\newblock \href {https://doi.org/10.1109/ROBOT.1997.620087}
  {\path{doi:10.1109/ROBOT.1997.620087}}.

\bibitem{chocron2008evolutionary}
O.~Chocron, Evolutionary design of modular robotic arms, Robotica 26~(3) (2008)
  323--330.
\newblock \href {https://doi.org/10.1017/s0263574707003931}
  {\path{doi:10.1017/s0263574707003931}}.

\bibitem{Icer2016}
M.~A. Esra~Icer, Andrea~Giusti, A task-driven algorithm for conﬁguration
  synthesis of modular robots, in: IEEE International Conference on Robotics
  and Automation (ICRA), 2016.
\newblock \href {https://doi.org/10.1109/icra.2016.7487727}
  {\path{doi:10.1109/icra.2016.7487727}}.

\bibitem{icer2017evolutionary}
E.~Icer, H.~A. Hassan, K.~El-Ayat, M.~Althoff, Evolutionary cost-optimal
  composition synthesis of modular robots considering a given task, in:
  Intelligent Robots and Systems (IROS), 2017 IEEE/RSJ International Conference
  on, IEEE, 2017, pp. 3562--3568.
\newblock \href {https://doi.org/10.1109/iros.2017.8206201}
  {\path{doi:10.1109/iros.2017.8206201}}.

\bibitem{moulianitis2016task}
V.~C. Moulianitis, A.~I. Synodinos, C.~D. Valsamos, N.~A. Aspragathos,
  Task-based optimal design of metamorphic service manipulators, Journal of
  Mechanisms and Robotics 8~(6) (2016) 061011.
\newblock \href {https://doi.org/10.1115/1.4033665}
  {\path{doi:10.1115/1.4033665}}.

\bibitem{brandstotter2015curved}
M.~Brandst{\"o}tter, A.~Angerer, M.~Hofbaur, The curved manipulator (cuma-type
  arm): Realization of a serial manipulator with general structure in modular
  design, in: Proceedings of the 14th IFToMM World Congress, 2015, pp.
  403--409.
\newblock \href {https://doi.org/10.6567/IFToMM.14TH.WC.OS2.037}
  {\path{doi:10.6567/IFToMM.14TH.WC.OS2.037}}.

\bibitem{dograUnified2022}
A.~Dogra, S.~Mahna, S.~S. Padhee, E.~Singla,
  \href{https://www.sciencedirect.com/science/article/pii/S0736584522000722}{Unified
  modeling of unconventional modular and reconfigurable manipulation system},
  Robotics and Computer-Integrated Manufacturing 78 (2022) 102385.
\newblock \href {https://doi.org/10.1016/j.rcim.2022.102385}
  {\path{doi:10.1016/j.rcim.2022.102385}}.
\newline\urlprefix\url{https://www.sciencedirect.com/science/article/pii/S0736584522000722}

\bibitem{dograRobotica2021}
A.~Dogra, S.~S. Padhee, E.~Singla, Optimal architecture planning of modules for
  reconfigurable manipulators, Robotica (2021) 1--15\href
  {https://doi.org/10.1017/S0263574720001174}
  {\path{doi:10.1017/S0263574720001174}}.

\bibitem{dograJMD2022}
A.~Dogra, S.~Sekhar~Padhee, E.~Singla,
  \href{https://doi.org/10.1115/1.4054336}{{Optimal Synthesis of Unconventional
  Links for Modular Reconfigurable Manipulators}}, Journal of Mechanical Design
  144~(8), 083304 (04 2022).
\newblock \href {https://doi.org/10.1115/1.4054336}
  {\path{doi:10.1115/1.4054336}}.
\newline\urlprefix\url{https://doi.org/10.1115/1.4054336}

\bibitem{kinova2019}
A.~Campeau-Lecours, H.~Lamontagne, S.~Latour, P.~Fauteux, V.~Maheu, F.~Boucher,
  C.~Deguire, L.-J.~C. L'Ecuyer, Kinova modular robot arms for service robotics
  applications, in: Rapid Automation: Concepts, Methodologies, Tools, and
  Applications, IGI Global, 2019, pp. 693--719.
\newblock \href {https://doi.org/10.4018/978-1-5225-8060-7.ch032}
  {\path{doi:10.4018/978-1-5225-8060-7.ch032}}.

\bibitem{2083}
E.~G. Gilbert, D.~W. Johnson, S.~S. Keerthi, A fast procedure for computing the
  distance between complex objects in three-dimensional space, IEEE Journal on
  Robotics and Automation 4~(2) (1988) 193--203.
\newblock \href {https://doi.org/10.1109/56.2083} {\path{doi:10.1109/56.2083}}.

\bibitem{kuffner2000rrt}
J.~J. Kuffner, S.~M. LaValle, Rrt-connect: An efficient approach to
  single-query path planning, in: Proceedings 2000 ICRA. Millennium Conference.
  IEEE International Conference on Robotics and Automation. Symposia
  Proceedings (Cat. No. 00CH37065), Vol.~2, IEEE, 2000, pp. 995--1001.
\newblock \href {https://doi.org/10.1109/ROBOT.2000.844730}
  {\path{doi:10.1109/ROBOT.2000.844730}}.

\end{thebibliography}

\end{document}